\documentclass{egpubl}

\usepackage{eg2026}
 
\ConferencePaper        

\CGFStandardLicense  

\usepackage[T1]{fontenc}
\makeatletter
\def\input@path{{EGauthor/}}
\makeatother
\usepackage{dfadobe}

\newcommand{\pname}[1]{{OUGS}} 
\usepackage[table]{xcolor}     
\definecolor{best}{rgb}{1, 0.7, 0.7}    
\definecolor{second}{rgb}{1, 0.85, 0.7}
\definecolor{third}{rgb}{1, 1, 0.7}
\definecolor{darkred}{RGB}{160, 0, 0} 

\newcommand{\rev}[1]{#1}

\usepackage{siunitx}
\usepackage{multirow}
\usepackage{booktabs}
\usepackage{diagbox}
\usepackage{subcaption}
\usepackage{amssymb} 
\usepackage{amsmath}

\usepackage{cite}  
\BibtexOrBiblatex
\electronicVersion
\PrintedOrElectronic

\ifpdf \usepackage[pdftex]{graphicx} \pdfcompresslevel=9
\else \usepackage[dvips]{graphicx} \fi

\usepackage{egweblnk} 

\title[Object-aware Uncertainty for 3DGS]%
      {OUGS: Active View Selection via Object-aware Uncertainty Estimation in 3DGS}

\author[H. Li, Q. Chen, D. Kalkofen, \& H.-T. Chen]
{\parbox{\textwidth}{\centering Haiyi Li$^{1}$\orcid{0009-0004-6914-8457},
         Qi Chen$^{1}$\orcid{0000-0001-8732-8049},
         Denis Kalkofen$^{2}$\orcid{0000-0002-0359-206X},
         and Hsiang-Ting Chen$^{1}$\orcid{0000-0003-0873-2698}
        }
        \\
{\parbox{\textwidth}{\centering 
         $^1$University of Adelaide, Australia \\
         $^2$Graz University of Technology
        }
}
}

\begin{document}


\maketitle
\begin{abstract}
Recent advances in 3D Gaussian Splatting (3DGS) have achieved state-of-the-art results for novel view synthesis. However, efficiently capturing high-fidelity reconstructions of specific objects within complex scenes remains a significant challenge. A key limitation of existing active reconstruction methods is their reliance on scene-level uncertainty metrics, which are often biased by irrelevant background clutter and lead to inefficient view selection for object-centric tasks.
We present \pname{}, a novel framework that addresses this challenge with a more principled, physically-grounded uncertainty formulation for 3DGS. Our core innovation is to derive uncertainty directly from the \textbf{explicit physical parameters} of the 3D Gaussian primitives (e.g., position, scale, rotation). By propagating the covariance of these parameters through the rendering Jacobian, we establish a highly interpretable uncertainty model. This foundation allows us to then seamlessly integrate semantic segmentation masks to produce a targeted, \textbf{object-aware} uncertainty score that effectively disentangles the object from its environment. This allows for a more effective active view selection strategy that prioritizes views critical to improving object fidelity.
Experimental evaluations on public datasets demonstrate that our approach significantly improves the efficiency of the 3DGS reconstruction process and achieves higher quality for targeted objects compared to existing state-of-the-art methods, while also serving as a robust uncertainty estimator for the global scene.


\end{abstract}
\section{Introduction}
\label{sec:intro}
Efficient 3D scene reconstruction is a foundational goal in computer vision and robotics. The advent of Neural Radiance Fields (NeRF)~\cite{mildenhall2020nerf} marked a breakthrough, enabling photorealistic novel view synthesis by learning implicit volumetric representations. More recently, 3D Gaussian Splatting (3DGS)~\cite{kerbl3DGaussianSplatting2023} has emerged as a compelling alternative. By modeling scenes with explicit 3D Gaussian primitives and leveraging a fast, differentiable rasterization pipeline, 3DGS achieves real-time rendering speeds without compromising visual fidelity, addressing the high computational cost that limits NeRF’s practicality.

Despite these advances, both NeRF and 3DGS remain highly data-intensive, typically requiring dense image captures to produce high-quality reconstructions~\cite{niemeyer2021regnerfregularizingneuralradiance, celarek2025does3dgaussiansplatting}. This motivates the need for active reconstruction~\cite{chen2011active}, where a human~\cite{shen2025educator,minton2025longitudinal} or robotic agent actively selects a minimal subset of views that maximally reduces uncertainty. Achieving this requires models to estimate their own information needs in real-time—a capability that hinges on accurate uncertainty estimation.

\begin{figure}[t!]
    \centering
    \vspace{-1mm}
    \includegraphics[width=\linewidth]{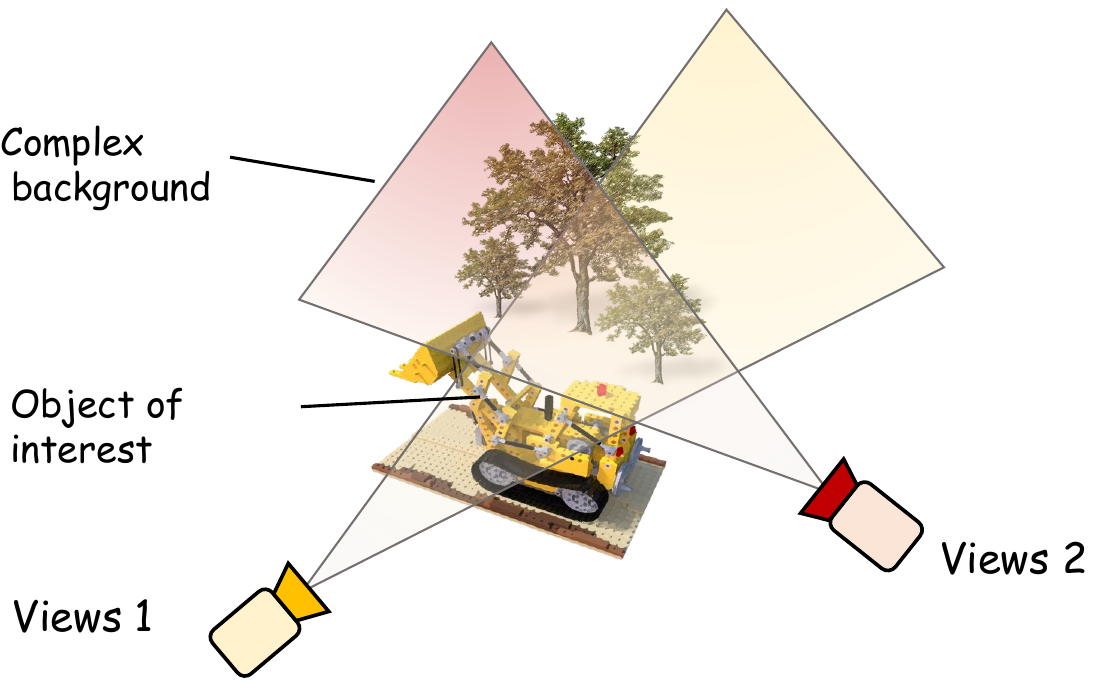}
    \caption{\textbf{A complex background can inflate image-level uncertainty.} \textit{This can mislead active view selection away from the object.}}
    \vspace{-7mm}
    \label{fig:obj_uncertainty}
\end{figure}

\begin{figure*}[t]
    \centering
    \includegraphics[width=0.9\linewidth]{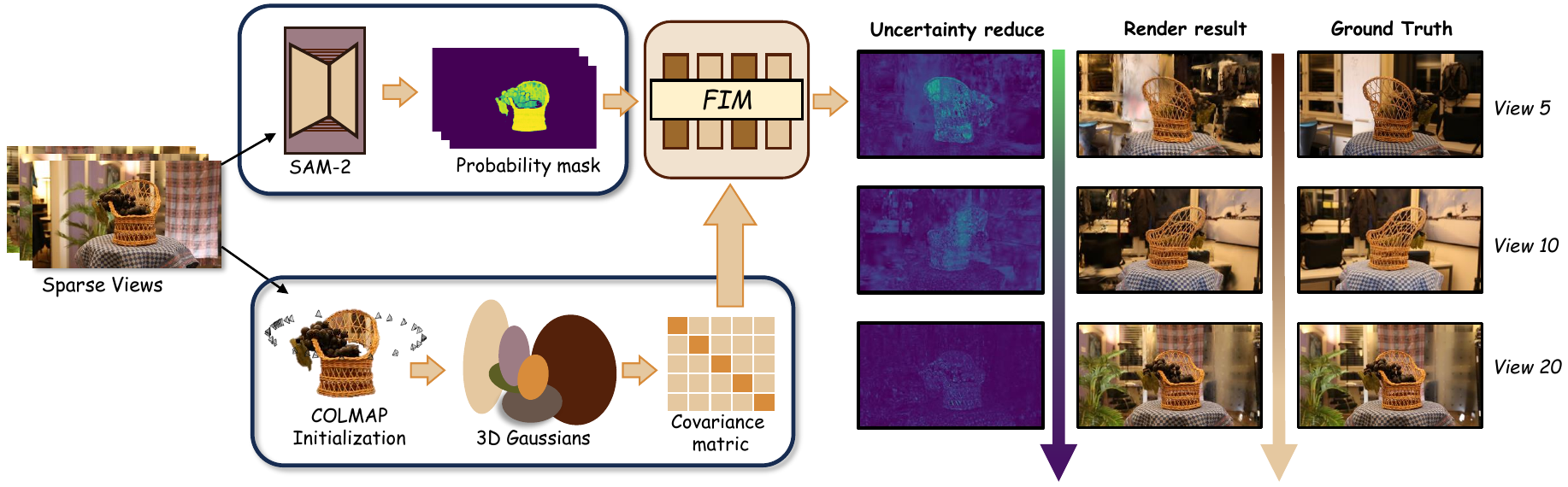}
    \caption{\textbf{Object‑aware uncertainty guides 3DGS view planning for precise object reconstruction.} \textit{Our physically-grounded uncertainty model, derived from the explicit parameters of the 3D Gaussians, is combined with a semantic mask to generate an object-level uncertainty score. This score effectively guides the active view selection to improve object fidelity, as shown for 5, 10, and 20 selected views.}}
    \label{fig:overview}
    \vspace{-5mm}
\end{figure*}

In this context, recent work has explored incorporating uncertainty into both NeRFs~\cite{goli2023,klassonSourcesUncertainty3D2024} and 3DGS. For the latter, emerging research has methods based on ensemble variance~\cite{hanViewDependentUncertaintyEstimation2025} or Fisher Information approximations~\cite{jiangFisherRFActiveView2024}. However, these pioneering methods share a fundamental limitation: they typically estimate uncertainty at the scene level. As illustrated in figure~\ref{fig:obj_uncertainty}, a global uncertainty score is often dominated by complex but irrelevant background clutter, misleading the view selection process. This is particularly problematic for applications where the primary goal is to capture a \textbf{specific object of interest} with the highest possible fidelity, such as in cultural heritage preservation, industrial inspection, and AR~\cite{cortes2023eeg,wu2023effect}. These scenarios demand object-centric planning, where background noise can distort the focus.

To address this critical gap, we introduce \pname{}, a framework designed specifically for \textbf{object-centric active reconstruction} that prioritizes the most informative camera viewpoints during training to improve reconstruction quality and efficiency (Figure~\ref{fig:overview})
Our work is built upon a key insight: to effectively isolate an object's uncertainty, one must first model uncertainty from a more fundamental, physically-grounded source, as is common practice, our method is the first to establish a rigorous framework that quantifies uncertainty directly from the \textbf{explicit physical parameters} of the 3D Gaussian primitives—their position, scale, rotation, and appearance. We begin by treating these parameters as random variables and propagate their covariance through the differentiable rendering pipeline via the Jacobian. This yields a pixel-wise visual uncertainty map that is not only robust but also highly interpretable.
This covariance captures epistemic uncertainty over existing Gaussian parameters—precisely what next-best-view selection aims to reduce during refinement. Since unobserved regions may lack instantiated Gaussians after sparse initialization, our method focuses on improving already-represented areas rather than discovery of entirely new structure.

This explicit projection to pixel space represents a crucial advantage over implicit methods like FisherRF~\cite{jiangFisherRFActiveView2024}. While FisherRF optimizes an information-theoretic objective in the abstract parameter space of a neural field, our approach generates a native spatial uncertainty field. This spatial modularity allows us to seamlessly integrate semantic masks as direct filters, effectively disentangling the target object's uncertainty from its environment.

To ensure scalability, we approximate the parameter covariance using a diagonal Fisher Information Matrix (FIM), updated efficiently through an exponential moving average. This complete formulation enables an active view selection strategy that is powerfully and precisely focused on the object of interest. Our contributions are threefold:
    \begin{itemize}
        \item We introduce a novel active reconstruction framework specifically designed for \textbf{object-centric tasks} in 3DGS, addressing a key limitation of existing scene-level methods.
        \item We propose a new, \textbf{physically-grounded uncertainty model} based on the explicit parameters of 3D Gaussians, offering greater accuracy and interpretability compared to implicit, weight-based approaches.
        \item Through extensive experiments, we demonstrate that our method significantly outperforms state-of-the-art approaches in object-focused reconstruction while maintaining strong performance on global scene metrics.
    \end{itemize}

\vspace{-2mm}
\section{Related Work}
\subsection{Uncertainty in 3D Splatting}
\label{sec:related-3dgs}

Quantifying uncertainty in 3DGS is an emerging research area crucial for real-world applications. Current approaches can be categorized into four main directions:
\textbf{(1) Variational/Bayesian.}
A principled approach is to treat Gaussian parameters as distributions. Li \& Cheung~\cite{NEURIPS2024_a076d0d1} use hierarchical Bayesian priors, while Savant et al.~\cite{wilsonModelingUncertainty3D2024} employ variational inference. While mathematically rigorous, these methods incur significant computational overhead, limiting their real-time applicability.
\textbf{(2) Sensitivity-based pruning.}
Alternatively, some methods measure the model's sensitivity to its parameters. PUP 3D-GS, for instance, uses a Hessian-based metric to prune Gaussians with high uncertainty. This approach is efficient but offers a less direct measure of predictive uncertainty.
\textbf{(3) Learned uncertainty fields.}
Several works train auxiliary predictors to output uncertainty maps or uncertainty-related signals.
UNG-GS~\cite{tanUncertaintyAwareNormalGuidedGaussian2025} introduces an uncertainty-aware field to better handle sparse inputs, while Han \& Dumery~\cite{hanViewDependentUncertaintyEstimation2025} learn a view-dependent uncertainty field for 3DGS.
These methods are flexible, but the resulting uncertainty is typically model-dependent and may be less physically interpretable than parameter-centric formulations.
We note that some concurrent 3DGS uncertainty systems (including UNG-GS at the time of our experiments) did not provide a publicly available implementation, which prevented us from reproducing their exact pipelines and performing controlled, apples-to-apples comparisons under our object-centric evaluation protocol.
\textbf{(4) Information-theoretic.}
Finally, information theory can be used to quantify the information value of additional views. GauSS-MI~\cite{xieGauSSMIGaussianSplatting2025}, for instance, selects views that maximize mutual information. While powerful for view selection, this paradigm focuses on the information value of candidate views rather than directly modeling the inherent uncertainty of the current reconstruction.

Our work carves a distinct path by adopting an efficient, FIM-based approximation of parameter uncertainty. We apply this formulation directly to \emph{object-centric active view selection}---a critical application gap not fully addressed by prior works.

\subsection{Uncertainty for Active View Selection}
\label{sec:related-nbv}

Active view selection, or Next-Best-View (NBV) planning, is a long-standing problem in computer vision and robotics~\cite{chen2011active}, aiming to intelligently choose views to maximize reconstruction quality while minimizing cost. Methodologies have evolved significantly over time.
\textbf{(1) Traditional and geometric methods.}
Early approaches in robotics often relied on geometric heuristics. For instance, receding-horizon planners like the one by Bircher et al.~\cite{bircherRecedingHorizonNextBestView2016} aim to maximize exploration of unknown free space using occupancy maps. Other classical NBV methods use voxel-grid representations and select views based on metrics like Shannon entropy or frontier exploration~\cite{kim2022infonerfrayentropyminimization}. While effective for coverage, these discretized methods can struggle to capture fine geometric details and are less suited for the continuous representations used in modern neural rendering.
\textbf{(2) Uncertainty- and information-driven neural rendering.}
With neural rendering, view selection can be guided by predictive uncertainty or information gain.
ActiveNeRF~\cite{panActiveNeRFLearningWhere2022} uses variance-driven criteria, while FisherRF~\cite{jiangFisherRFActiveView2024} proposes Fisher-information-based objectives for principled NBV selection.
Concurrent works extend these ideas to 3DGS: POp-GS~\cite{wilsonPOpGSNextBest2025a} uses Fisher-matrix-based planning, while GauSS-MI~\cite{xieGauSSMIGaussianSplatting2025} ranks views by mutual information. However, these methods optimize \emph{global} scene-level objectives, where background clutter can dominate the planning signal—suboptimal when reconstructing a specific object. As POp-GS lacked public code during our study, we compare against baselines with available implementations under controlled settings, using identical masks where applicable.
\textbf{(3) Other view-selection paradigms.}
Beyond uncertainty and information-theoretic approaches, other paradigms have been explored. Learning-based methods, such as NeurAR~\cite{NeurARRan_2023}, employ reinforcement learning to train an agent that learns an optimal view selection policy directly from simulation. Concurrently, other works focus on explicitly modeling visibility. For instance, Neural Visibility Fields~\cite{xue2024neuralvisibilityfielduncertaintydriven} learn to predict which parts of a scene are visible from a given viewpoint, guiding selection towards views that maximize observable new area. While powerful, these methods either require extensive training or shift the focus from reconstruction fidelity to geometric coverage.

Our work addresses the critical limitation of scene-level uncertainty methods. By introducing an object-aware mechanism built on a physically grounded, parameter-centric uncertainty formulation, we enable the view selection process to focus on semantically important regions of the scene, a challenge not explicitly addressed by any of these prior paradigms.

\vspace{-3mm}
\section{Method}
\label{sec:method}

\subsection{Preliminary: 3D Gaussian Splatting}
\label{sec:gaussian-rendering}

\begin{figure}[ht]
    \centering
    \includegraphics[width=\linewidth]{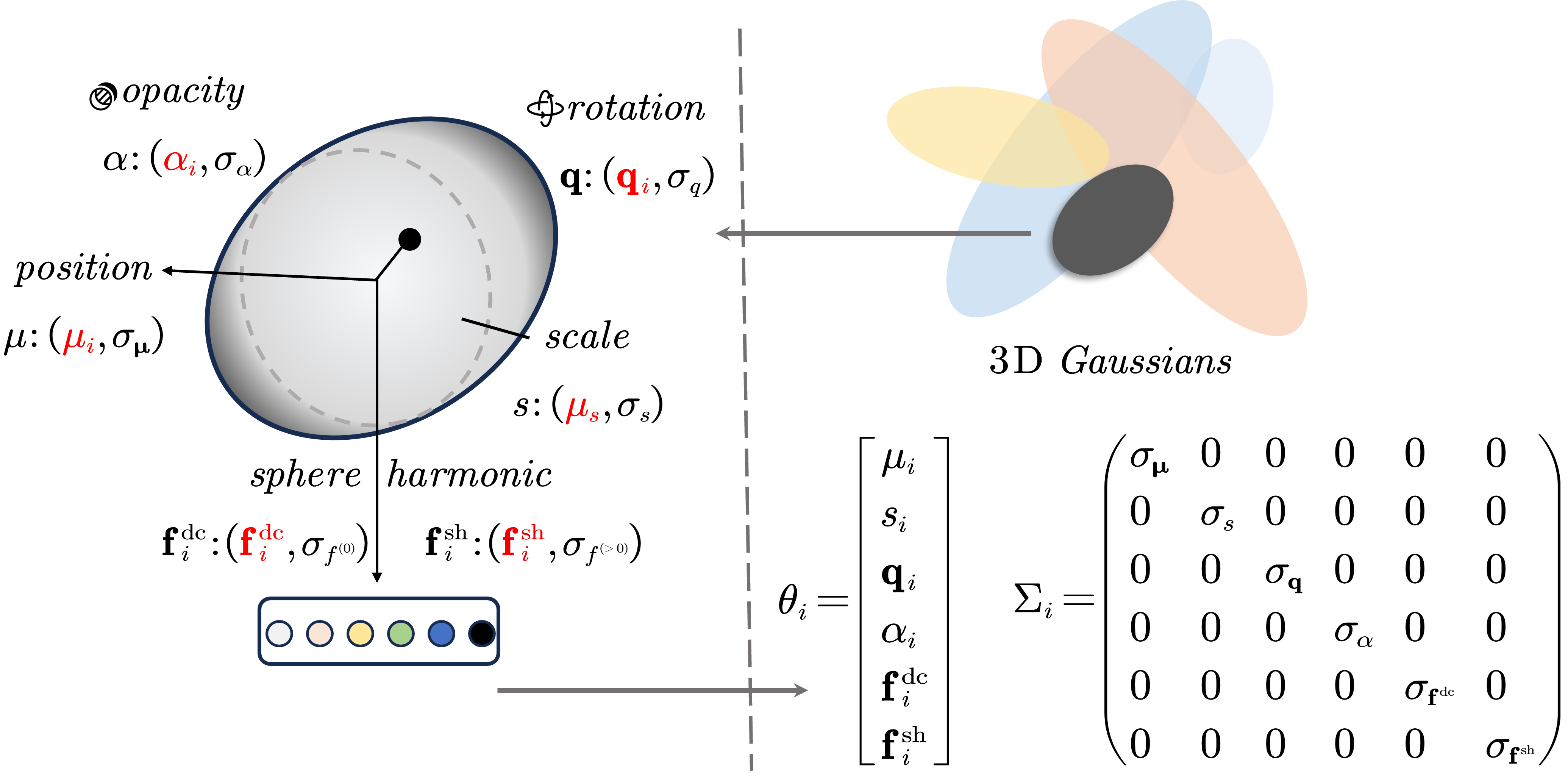}
\caption{\textbf{The parameterization of a 3D Gaussian primitive.} \textit{Each Gaussian, the fundamental building block of our scene representation, is defined by a set of explicit physical parameters.}}
    \label{fig:3D_Gaussian_representation}
\vspace{-4mm}
\end{figure}

Our method builds upon the 3D Gaussian Splatting (3DGS) framework~\cite{kerbl3DGaussianSplatting2023}, which represents a scene as a collection of anisotropic 3D Gaussian primitives $\mathcal{G} = \{ \mathcal{G}_i \}_{i=1}^{N_g}$. To ground our uncertainty analysis, we first provide a detailed list of the parameters used in the differentiable rendering pipeline. Each Gaussian $\mathcal{G}_i$ is fully described by a parameter vector $\theta_i$:
\begin{equation}
\theta_i = \left[
\underbrace{\boldsymbol{\mu}_i}_{\text{Center}},
\underbrace{\mathbf{s}_i}_{\text{Scale}},
\underbrace{\mathbf{q}_i}_{\text{Rotation}},
\underbrace{\alpha_i}_{\text{Opacity}},
\underbrace{\mathbf{f}_i^{\text{dc}}, \mathbf{f}_i^{\text{sh}}}_{\text{Color (SH)}}
\right]^\top \label{eq:gaussian_params}
\end{equation}

\noindent where the components are:
\begin{itemize}
    \item Geometry: The 3D center $\boldsymbol{\mu}_i \in \mathbb{R}^3$, an anisotropic scaling vector $\mathbf{s}_i \in \mathbb{R}^3_+$, and an orientation quaternion $\mathbf{q}_i \in \mathbb{S}^3$. Together, these define the Gaussian's position, size, and orientation.
    \item Appearance: A scalar opacity value $\alpha_i \in \mathbb{R}$ and view-dependent color modeled by Spherical Harmonics (SH). The color is parameterized by the degree-0 (DC) term $\mathbf{f}_i^{\text{dc}} \in \mathbb{R}^3$ and a set of higher-order coefficients $\mathbf{f}_i^{\text{sh}} \in \mathbb{R}^{3 \times 15}$.
\end{itemize}
A visual breakdown of these parameters is provided in figure~\ref{fig:3D_Gaussian_representation}.

Rendering in 3DGS uses a differentiable splatting approach based on standard alpha compositing. First, the 3D Gaussians are projected onto the 2D image plane and sorted in front-to-back order based on their depth. The color $C(u)$ for a pixel $u$ is then:
\begin{equation}
C(u) = \sum_{i \in \mathcal{I}(u)} c_i(u) \alpha'_i(u) \prod_{j=1}^{i-1} \left(1 - \alpha'_j(u)\right)
\end{equation}
where $\mathcal{I}(u)$ is the list of Gaussians overlapping the pixel $u$, $c_i(u)$ is the view-dependent color evaluated from SH. The effective opacity $\alpha'_i(u)$ is determined by modulating the Gaussian's learned opacity parameter $\alpha_i$ by its projected 2D profile at the pixel location.

\vspace{-2mm}
\subsection{Mapping 3D Gaussian Parameter Uncertainty to Pixel-wise Object-aware Uncertainty}
\label{parameter-uncertainty}
To quantify uncertainty, we treat the parameter vector of each Gaussian as a random variable and initialize it with a Gaussian prior,
\(\theta_i \sim \mathcal{N}(\theta_i^0, \Sigma_i^0)\).
For notational clarity, we stack the per-Gaussian covariances into a block-diagonal matrix
\begin{equation}
  \Sigma = \operatorname{diag}\!\bigl(\Sigma_1,\dots,\Sigma_{N_g}\bigr) \in \mathbb{R}^{d\times d}    
\end{equation}
where \(d = N_g\cdot d_g\).
Throughout the rest of this section, \(\Sigma\) refers to this global covariance, while \(\Sigma_i\) denotes its \(i\)-th block. We then project the uncertainty of the 3DGS parameters into pixel space and describe how it interacts with a soft object mask.
Here, we present simplified expressions; full derivations and a second-order error bound are provided in the section A of the Supplementary Material.

\textbf{Decomposition of Uncertainty Sources.}
A key advantage of our explicit, parameter-centric formulation is the ability to decompose the total uncertainty into its underlying physical sources. We can partition the Gaussian parameter vector $\theta_i$ into a \textbf{geometry} component, $\theta_i^g = [\boldsymbol{\mu}_i, \mathbf{s}_i, \mathbf{q}_i]^\top$, and an \textbf{appearance} component, $\theta_i^a = [\alpha_i, \mathbf{f}_i^{\text{dc}}, \mathbf{f}_i^{\text{sh}}]^\top$. Consequently, the total parameter vector $\theta$ and its Jacobian $J_u$ can be similarly partitioned:
\begin{equation}
    \theta = \begin{bmatrix} \theta^g \\ \theta^a \end{bmatrix}, \quad J_u = \begin{bmatrix} J_u^g & J_u^a \end{bmatrix}
    \label{eq:decomposition}
\end{equation}
where $J_u^g = \partial C(u) / \partial \theta^g$ and $J_u^a = \partial C(u) / \partial \theta^a$. Assuming independence between geometric and appearance parameters (a reasonable simplification enforced by our diagonal FIM approximation), the pixel-wise color covariance from Eq.~\ref{eq:pixel-cov} can be expressed as a sum of two distinct sources:
\begin{equation}
    \Sigma_{C}(u) \approx \underbrace{J_u^g \Sigma^g (J_u^g)^{\top}}_{\text{Geometric Uncertainty}} + \underbrace{J_u^a \Sigma^a (J_u^a)^{\top}}_{\text{Appearance Uncertainty}}
    \label{eq:cov_decomposition}
\end{equation}
This decomposition is theoretically significant. It allows us to differentiate between uncertainty arising from poorly constrained object geometry (e.g., ambiguous boundaries, fine structures) and uncertainty from poorly observed appearance (e.g., complex materials, view-dependent effects). Implicit methods like FisherRF, which operate on abstract network weights, lack this inherent interpretability.

\textbf{Pixel-wise uncertainty} Assume complete set of scene parameters $\theta = \{\theta_i\}$ and $\theta^\star$ is the MAP estimate after optimization. 
For small parameter perturbations $\delta\theta = \theta - \theta^\star$, the change in pixel color $\delta C(u)$ can be linearly approximated using a first-order Taylor expansion:
\[
\delta C(u) \approx J_u \delta\theta .
\]
where the Jacobian $J_u=\partial C(u;\theta)/\partial\theta \in\mathbb R^{3\times d}$. 
Given $\mathbb E[\delta\theta]=0$ under a prior normal distribution, the induced pixel‑colour covariance can be written as 
\begin{equation}
\Sigma_{C}(u)=\operatorname{Var}[C(u;\theta)]
\;\approx\;J_u\,\Sigma\,J_u^{\!\top}
\label{eq:pixel-cov}
\end{equation}
where $\Sigma$ is the full parameter covariance. Eq~\ref{eq:pixel-cov} shows that the Jacobian acts as a lever arm that magnifies (or attenuates) each parameter’s uncertainty in proportion to that parameter’s influence on the pixel.

\textbf{Pixel-wise object-aware uncertainty} To estimate the uncertainty of a specific object $k$, we introduce a soft mask $M_k(u) \in [0, 1]$ based on semantic probabilities. This allows us to define an object-specific pixel covariance $\Sigma_{C,k}(u)$ by masking the standard error propagation formula:
\begin{equation}
\Sigma_{C,k}(u) \;=\; \bigl(M_k(u)\bigr)^{\!2}\;\bigl(J_u\,\Sigma\,J_u^{\!\top}\bigr)
\label{eq:mask}
\end{equation}

The mask term $M_k(u)$ is squared because covariance propagates quadratically via the Jacobian and its transpose.
\begin{figure*}[!t]
    \centering
    \includegraphics[width=\linewidth]{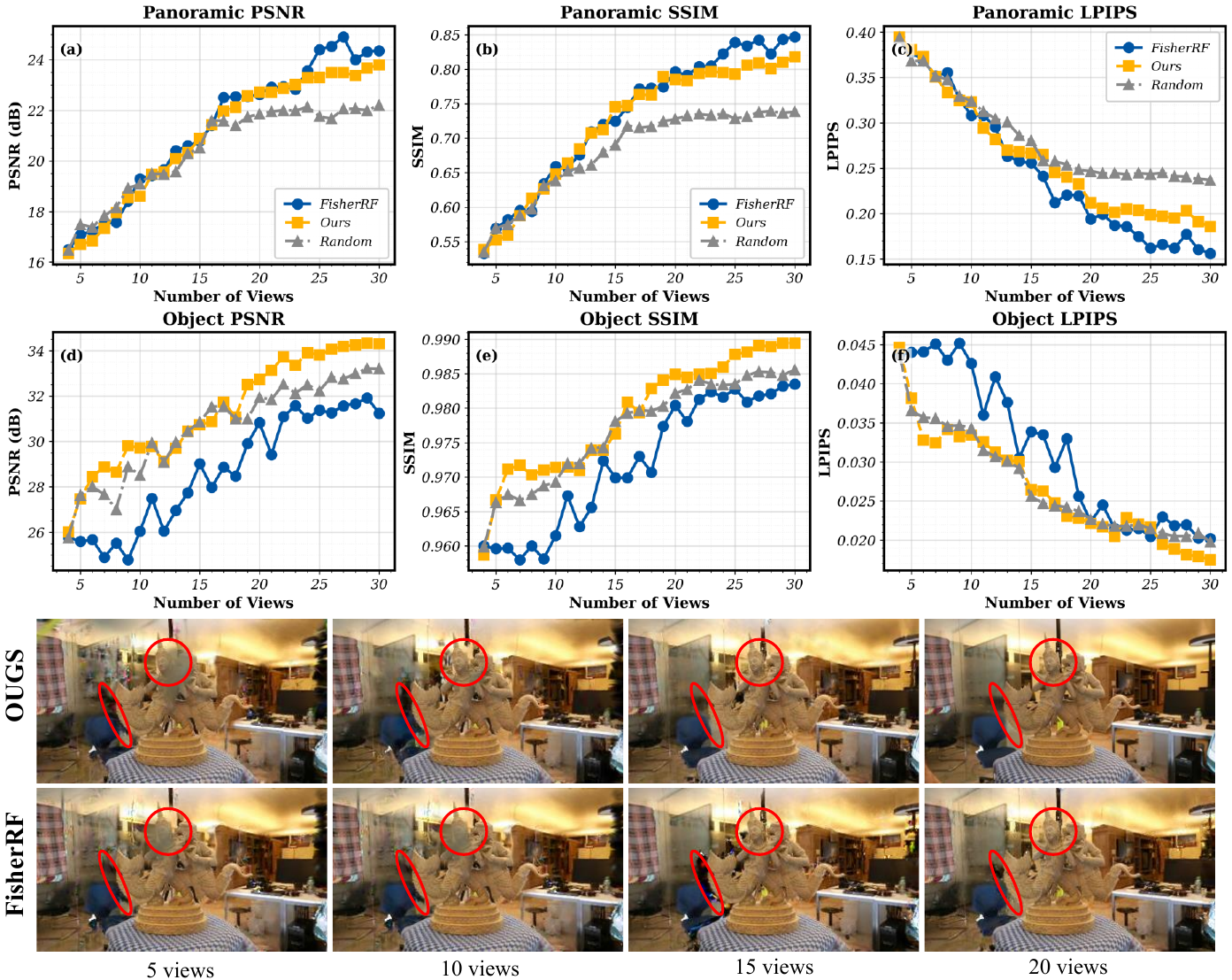}
    \caption{\textbf{Object‑aware approach speeds up convergence.}  
    \textit{Curves are recorded on the statue scene of the LF dataset as new views are added.  
    Top: panoramic PSNR/SSIM/LPIPS; middle: the same metrics evaluated only inside the object.  
    Bottom: visual progression at 5, 10, 15 and 20 views; red circles mark regions where the competing method keeps struggling while our reconstruction sharpens steadily.}}
    \label{fig:obj_active_view}
    \vspace{-4mm}
\end{figure*}
\vspace{-2mm}
\subsection{Updating Uncertainty With FIM}
\label{sec:fisher-ema}

While our formulation provides a physically interpretable model of uncertainty, direct computation of the full covariance matrix $\Sigma$ is intractable. We therefore approximate it with the inverse of the Fisher Information Matrix (FIM), $\Sigma \approx \sigma^{2}\mathcal{I}^{-1}$~\cite{FisherTutorial}. Crucially, our FIM is defined over the space of the 3D Gaussians' physical parameters, capturing how perturbations in geometry and appearance affect the rendered output. This stands in contrast to implicit-representation methods where the FIM is computed over abstract neural network weights.

\textbf{Online Diagonal Approximation.}
To ensure computational tractability, we make a key simplifying assumption: we approximate the full FIM with its diagonal entries only, effectively assuming that the different physical parameters of a Gaussian are locally independent. This diagonal approximation, $\mathcal{I} \approx \mathrm{diag}(\mathcal{I})$, aligns perfectly with the uncertainty decomposition presented in Eq.~\ref{eq:cov_decomposition}. It implies that a Gaussian's geometric uncertainty is decoupled from its appearance uncertainty. While this is a strong simplification, it is a common and effective strategy that allows us to efficiently estimate the parameter-wise variances. We update the diagonal entries $\mathcal{I}_{t,i}$ online during training using an exponential moving average (EMA) of squared gradients~\cite{kingma2017adammethodstochasticoptimization}:
\begin{equation}
\mathcal{I}_{t,i} = \alpha_t\,\mathcal{I}_{t-1,i} + (1 - \alpha_t)\,\left[\nabla_{\theta_i} \ell_t\right]^2,
\label{eq:fim_update}
\end{equation}
where $\ell_t$ denotes the photometric reconstruction loss at iteration $t$,
$\ell_t=\frac{1}{2\sigma^2}\sum_{u\in\Omega_t}\|C_{\text{pred}}(u)-C_{\text{gt}}(u)\|_2^2$ computed on the sampled pixels (or patches) at step $t$, and $[\cdot]^2$ is element-wise.
We employ a time-dependent momentum schedule $\alpha_t = 0.95^{\,1.5\,t/T}$.
This schedule gradually decreases the momentum over iterations (from $1$ at $t{=}0$ to $\approx 0.926$ at $t{=}T$), increasing the contribution of newly observed gradients as optimization stabilizes.

\textbf{Object-Aware Uncertainty Score.}
Substituting the diagonal approximation into Eq.~\ref{eq:mask} yields:
\begin{equation}
\Sigma_{C,k}(u) = \bigl(M_k(u)\bigr)^{2}\,J_u \bigl(\mathrm{diag}\,\mathcal{I}_{t} + \lambda I\bigr)^{-1} J_u^{\!\top},
\label{eq:final_uncertainty}
\end{equation}
where $\lambda$ is a small damping factor to ensure numerical stability.

Note that unlike expected information gain objectives that quantify parameter-space entropy reduction, our score sums pixel-space predictive variances, enabling direct spatial modulation via object masks. Key differences from FisherRF~\cite{jiangFisherRFActiveView2024}: (1) we project parameter covariance into pixel space, allowing native 2D mask integration for object-centric planning; (2) we maintain an online EMA estimate during training rather than snapshot-based gradient evaluation, stabilizing uncertainty under stochastic mini-batches.
\vspace{-2mm}

\begin{table*}[h]
\centering
\caption{\textbf{Active View Selection on Mip-NeRF360, Tanks\&Temples, and LF datasets}. \textit{"Panoramic" evaluates the full image; "Object-aware" evaluates only inside the object mask.
Rows denote selection policies.}}
\label{tab:full}
\resizebox{1\textwidth}{!}{
\small
\begin{tabular}{c|ccc|ccc|ccc}
    \noalign{\vskip .1em}        
    \noalign{\hrule height .2em} 
    \noalign{\vskip .1em}     
    \multirow{2}{*}{\diagbox{Method}{Metrics}} & \multicolumn{3}{c}{Mip-NeRF360} & \multicolumn{3}{c}{LF} & \multicolumn{3}{c}{Tanks \& Temples} \\
    & PSNR $\uparrow$ & SSIM $\uparrow$ & LPIPS $\downarrow$ & PSNR $\uparrow$ & SSIM $\uparrow$ & LPIPS $\downarrow$ & PSNR $\uparrow$ & SSIM $\uparrow$ & LPIPS $\downarrow$ \\
    \hline
    Random          & 17.9140 & 0.5640 & 0.4300 & 19.0857 & 0.6646 & 0.2669 & 15.8784 & 0.5365 & 0.3720\\
    ActiveNeRF       & 
    17.8890 & 0.5330 & 0.4140 & 21.2263 & 0.7691 & 0.1742 & 16.2918 & 0.5892 & \cellcolor{third}0.2514 \\
    BayesRays       & 18.8120 
    & 0.5730 & 0.4210 & 21.9232 & 0.7628 & 0.1752 & 16.9322 & 0.6091 & 0.3252 \\
    FisherRF      & 
    \cellcolor{third}20.3510 & \cellcolor{third}0.6010 & \cellcolor{third}0.3610 & \cellcolor{third}23.6450 & \cellcolor{second}0.8323 & \cellcolor{second}0.1651 & \cellcolor{second}17.3684 & \cellcolor{second}0.6296 & 0.3091\\
    GauSS-MI      & 
    \cellcolor{best}20.8150 & \cellcolor{second}0.6433 & \cellcolor{best}0.2710 & \cellcolor{best}23.9820 & \cellcolor{best}0.8354 & \cellcolor{best}0.1628 & \cellcolor{best}17.4210 & \cellcolor{best}0.6315 & \cellcolor{best}0.2425\\
    \pname{}            & \cellcolor{second}20.6099 & \cellcolor{best}0.6453 & \cellcolor{second}0.2727 & \cellcolor{second}23.7014 & \cellcolor{third}0.8058 & \cellcolor{third}0.1726 & \cellcolor{third}17.2666 & \cellcolor{third}0.6125 & \cellcolor{second}0.2468\\
    \noalign{\vskip .1em}        
    \noalign{\hrule height .15em} 
    \noalign{\vskip .1em}     
    \noalign{\vskip .1em}        
    \noalign{\hrule height .15em} 
    \noalign{\vskip .1em}     
    \multicolumn{10}{c}{\textbf{Object-Aware}} \\
    Random          & 26.3382 & 0.9732 & \cellcolor{third}0.0266 & \cellcolor{second}31.0709 & \cellcolor{second}0.9866 & \cellcolor{third}0.0227 & \cellcolor{third}24.6778 & \cellcolor{third}0.9306 & \cellcolor{third}0.1349\\
    
    FisherRF        & \cellcolor{third}26.4312 & \cellcolor{third}0.9731 & 0.0276 & 30.82740 & 0.9813 & 0.0231 & 23.4830 & 0.9208 & 0.1396\\
    GauSS-MI         & \cellcolor{second}27.3012 & \cellcolor{second}0.9764 & \cellcolor{second}0.0258 & \cellcolor{third}30.9742 & \cellcolor{third}0.9830 & \cellcolor{second}0.0226 & \cellcolor{second}24.9832 & \cellcolor{second}0.9318 & \cellcolor{second}0.1336\\
    \pname{}            & \cellcolor{best}29.6099 & \cellcolor{best}0.9813 & \cellcolor{best}0.0241 & \cellcolor{best}32.1856 & \cellcolor{best}0.9888 & \cellcolor{best}0.0221 & \cellcolor{best}26.2533 & \cellcolor{best}0.9333 & \cellcolor{best}0.1169\\
    \noalign{\vskip .1em}        
    \noalign{\hrule height .2em} 
    \noalign{\vskip .1em}     

\end{tabular}}
\vspace{-6mm}
\end{table*}

\section{Experiments}

\subsection{Experimental Setup}
\noindent\textbf{Dataset.} \label{sec:Dataset}
We conduct our evaluation across three public datasets to ensure coverage of diverse capture scenarios.
\textbf{Mip-NeRF 360}~\cite{barron2022mipnerf360unboundedantialiased} comprises four bounded indoor scenes with strong specular clutter and five unbounded outdoor environments featuring foliage occlusion and high dynamic range lighting.
Following the standard 3DGS protocol~\cite{kerbl3DGaussianSplatting2023}, every $8^{\text{th}}$ view is held out for testing.
We also evaluate on the \textbf{Light-Field} dataset~\cite{10.1145/2876504}, which features four tabletop objects (\emph{torch}, \emph{statue}, \emph{basket}, and \emph{africa}) captured via a motorised gantry.
A salient characteristic of this dataset is that the target object remains centred across all views. This yields stable masks from SAM-2, making it an ideal testbed for evaluating per-object uncertainty calibration.
Additionally, we select the \emph{train} and \emph{truck} scenes from \textbf{Tanks \& Temples}~\cite{tank}. These large-scale, drone-style outdoor captures are characterised by long-baseline parallax and strong depth discontinuities.
Inspired by the sparse-view protocol of Shen~\cite{variationfor3dreconstruction}, we adopt an intentionally biased and imbalanced camera orbit configuration. Such a setup exacerbates reconstruction artefacts, serving as a rigorous stress test for our Fisher-based uncertainty modelling. To preserve fine geometric cues, all RGB frames are processed at their full original resolution without cropping.
\begin{figure*}[!th]
  \centering
  \setlength{\tabcolsep}{0pt}      
  \setlength{\extrarowheight}{0pt} 
  \renewcommand{\arraystretch}{0}  
  \setlength{\aboverulesep}{0pt}
  \setlength{\belowrulesep}{0pt}
\scalebox{1}{%
\begin{tabular}{@{}
        *{4}{>{\centering\arraybackslash}m{0.24\linewidth}} 
        @{}}

\textbf{GT} & \textbf{\pname{}} & \textbf{FisherRF} & \textbf{Random} \\

\includegraphics[width=\linewidth]{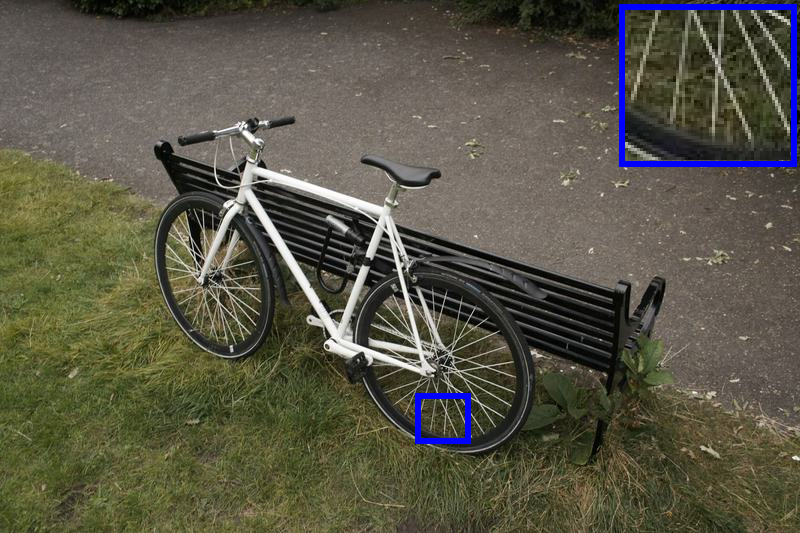} &
\includegraphics[width=\linewidth]{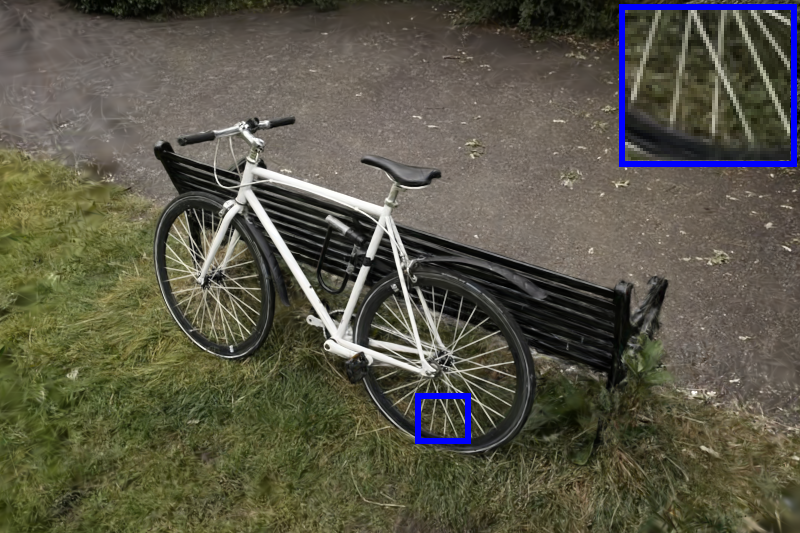} &
\includegraphics[width=\linewidth]{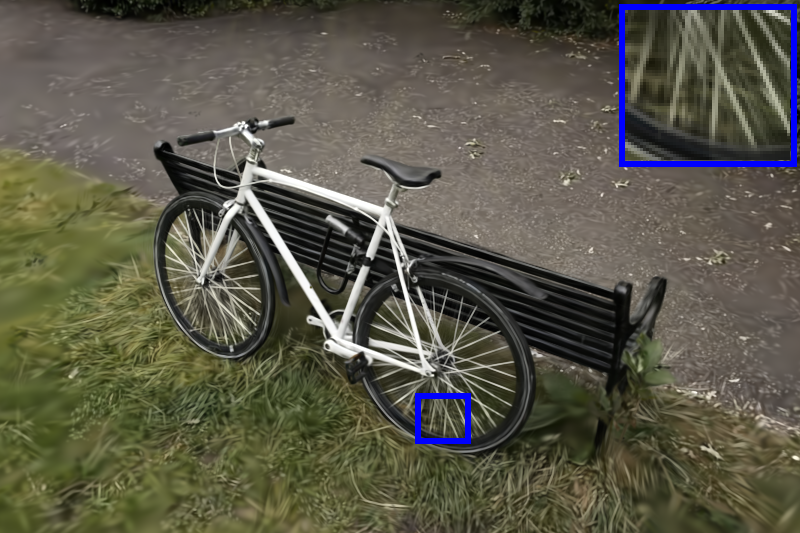} &
\includegraphics[width=\linewidth]{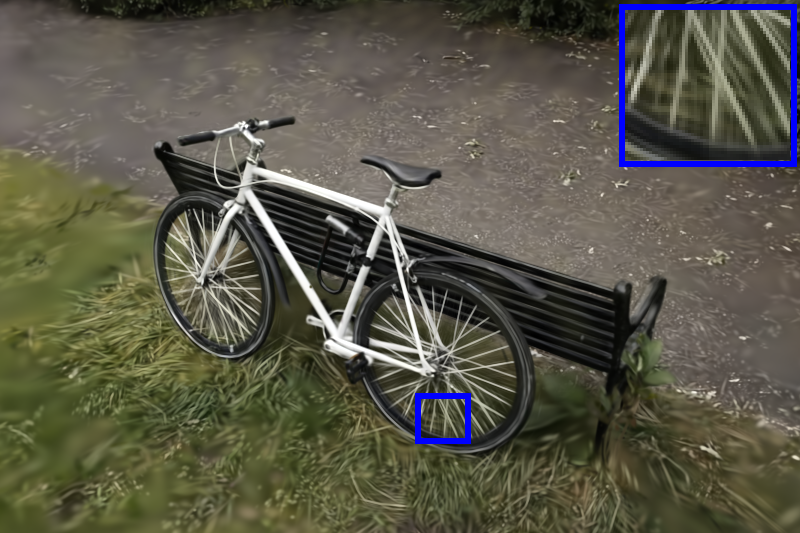} \\

\includegraphics[width=\linewidth]{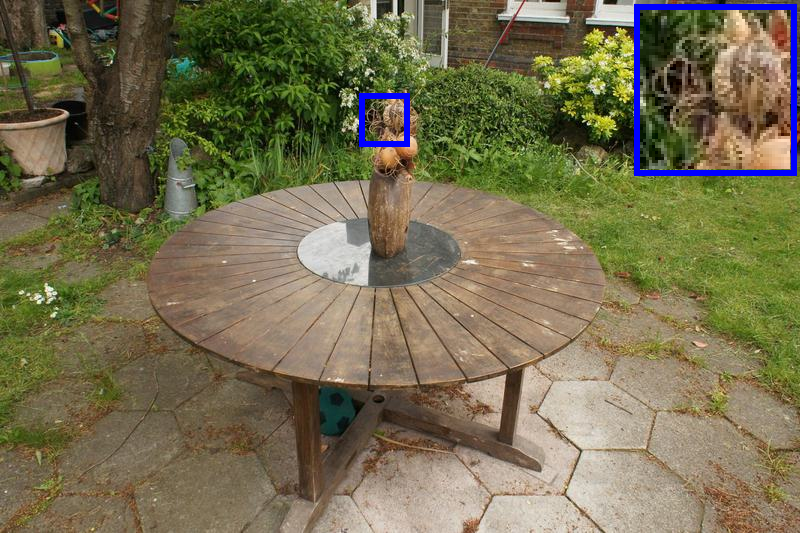} &
\includegraphics[width=\linewidth]{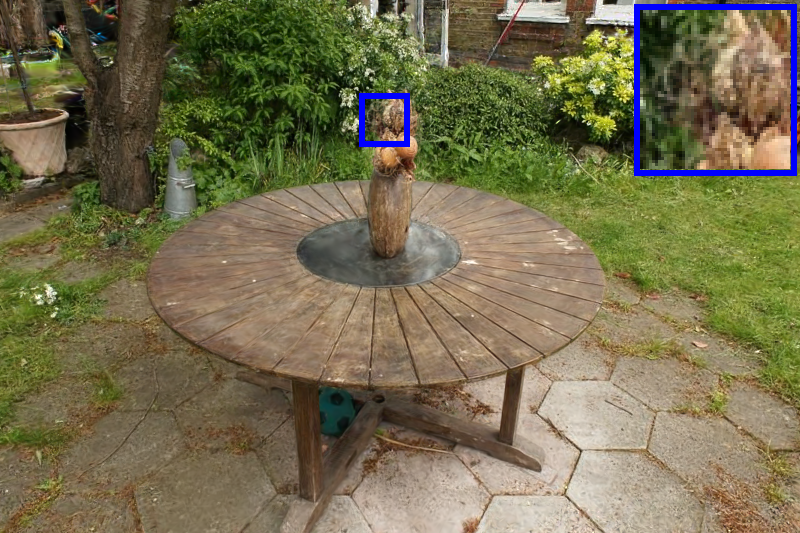} &
\includegraphics[width=\linewidth]{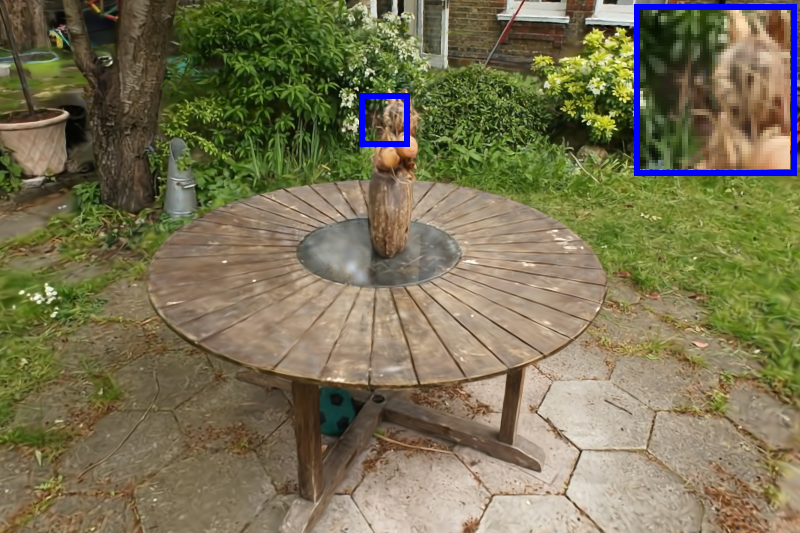} &
\includegraphics[width=\linewidth]{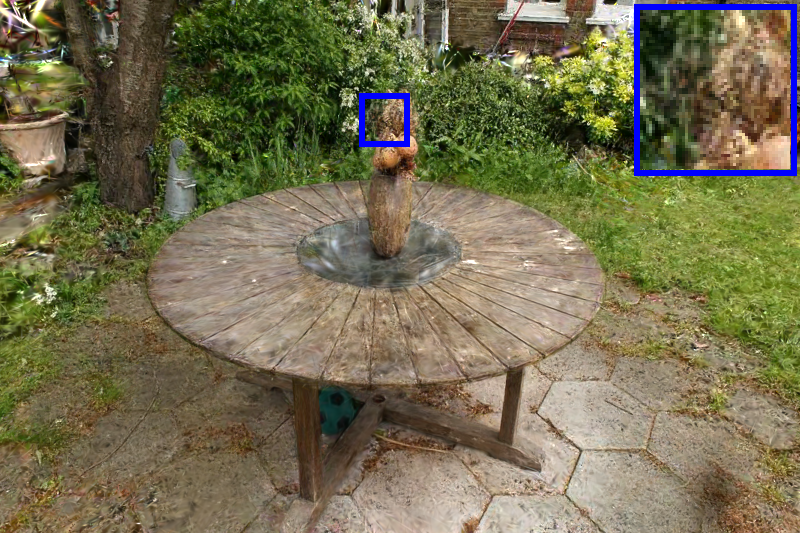} \\

\includegraphics[width=\linewidth]{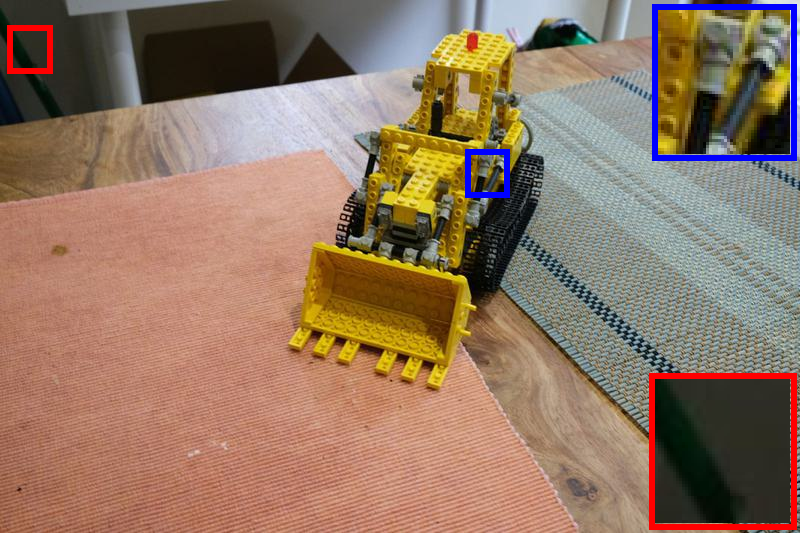} &
\includegraphics[width=\linewidth]{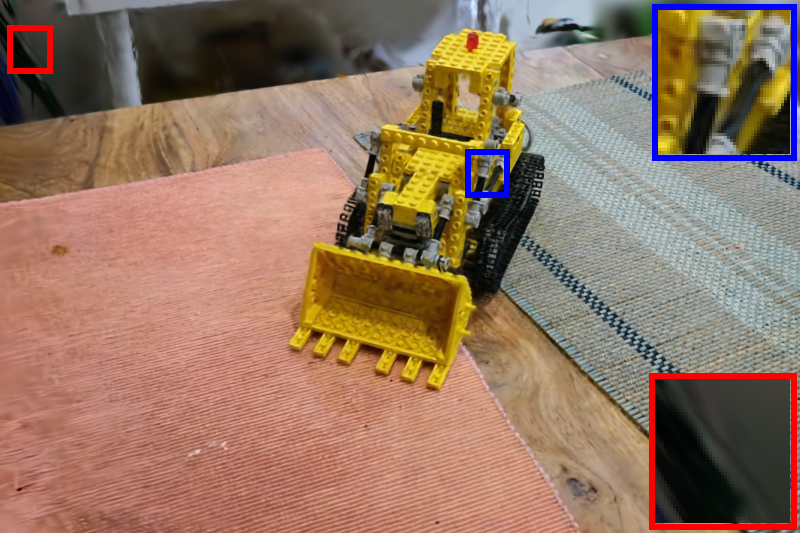} &
\includegraphics[width=\linewidth]{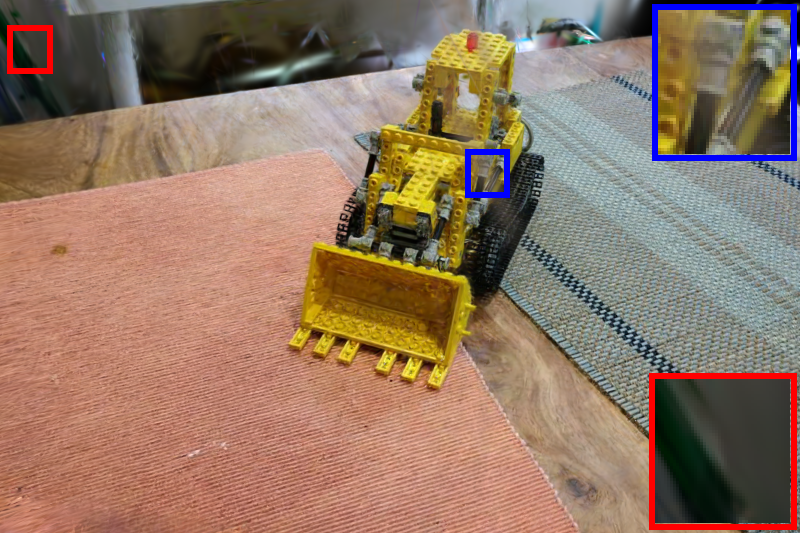} &
\includegraphics[width=\linewidth]{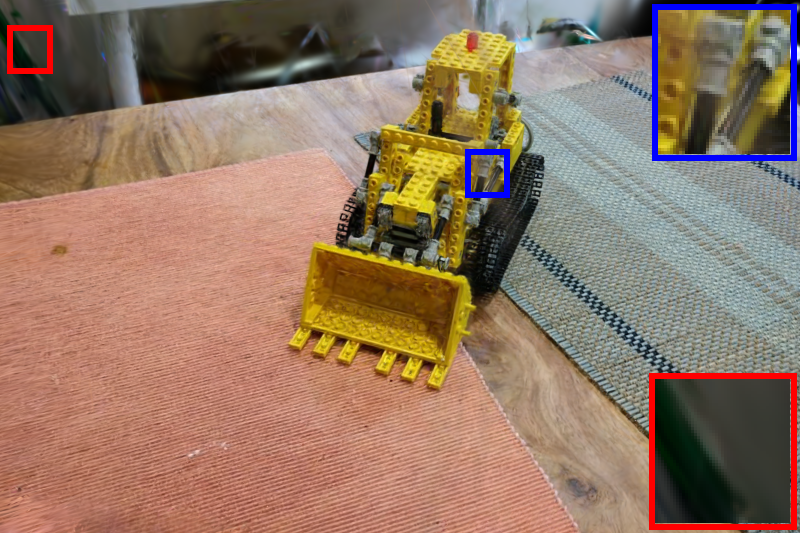} \\

\includegraphics[width=\linewidth]{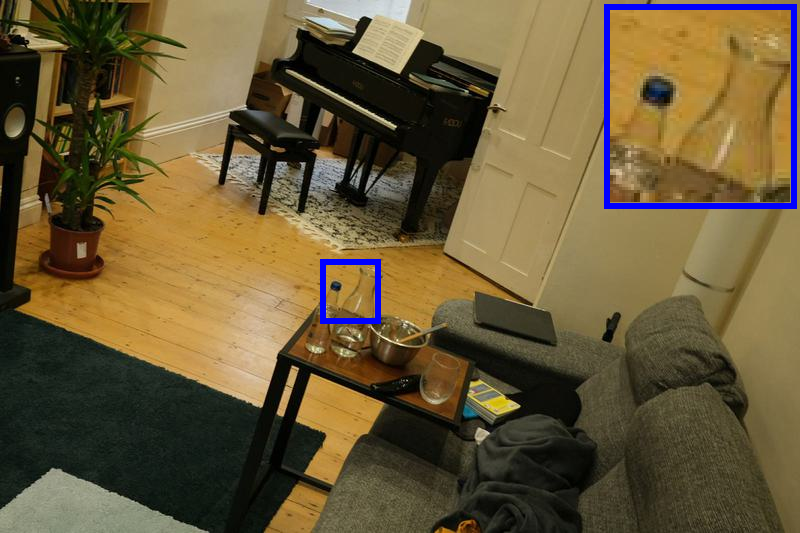} &
\includegraphics[width=\linewidth]{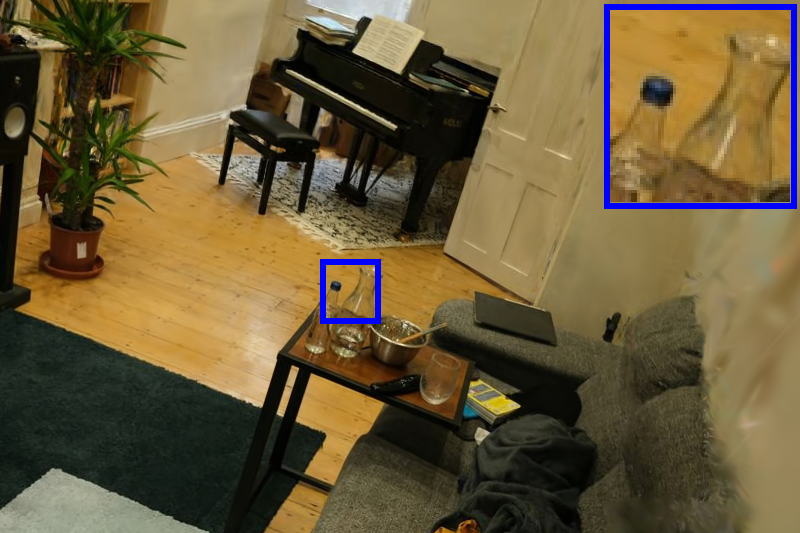} &
\includegraphics[width=\linewidth]{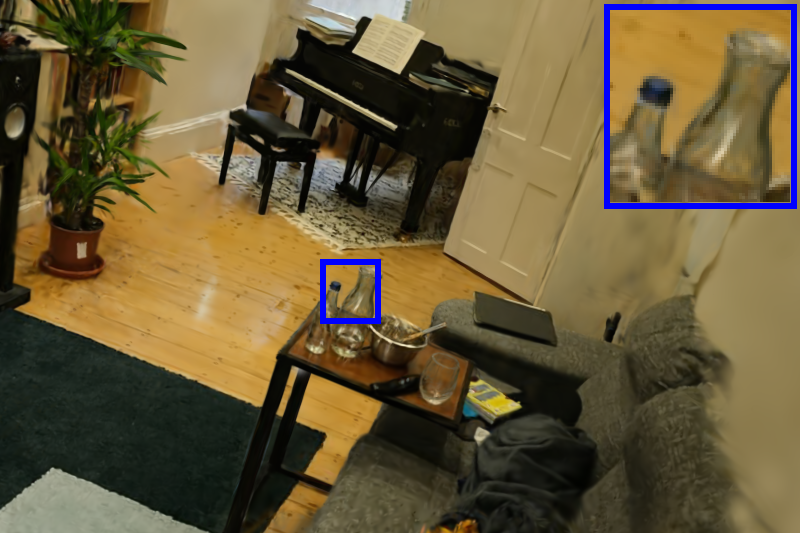} &
\includegraphics[width=\linewidth]{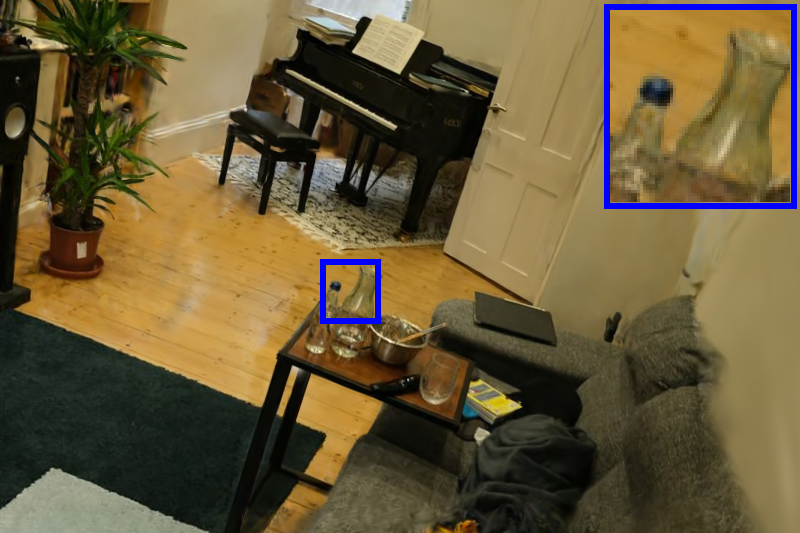} \\

\includegraphics[width=\linewidth]{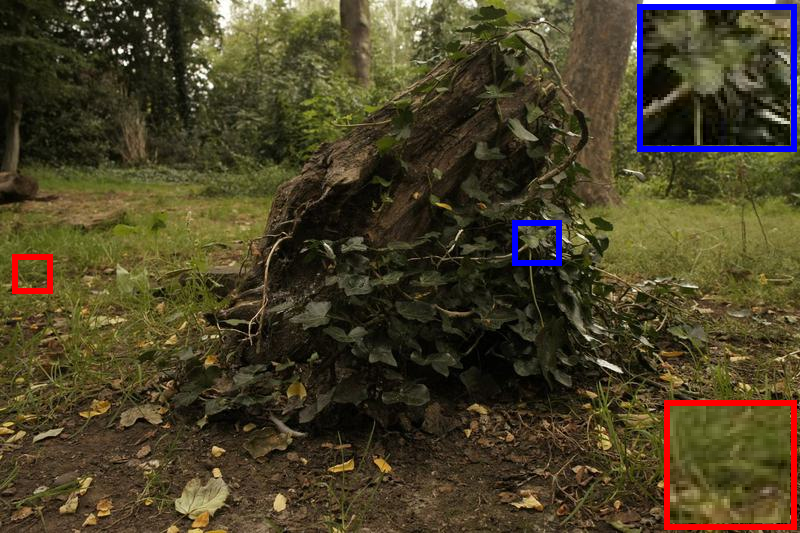} &
\includegraphics[width=\linewidth]{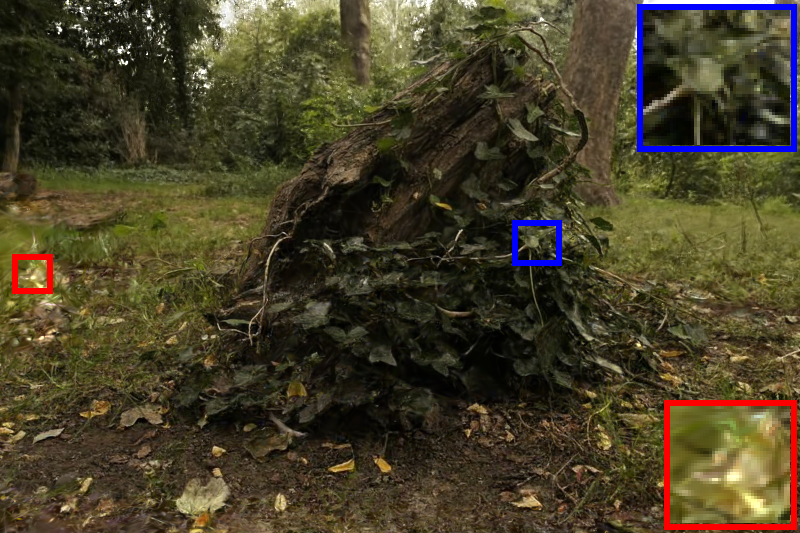} &
\includegraphics[width=\linewidth]{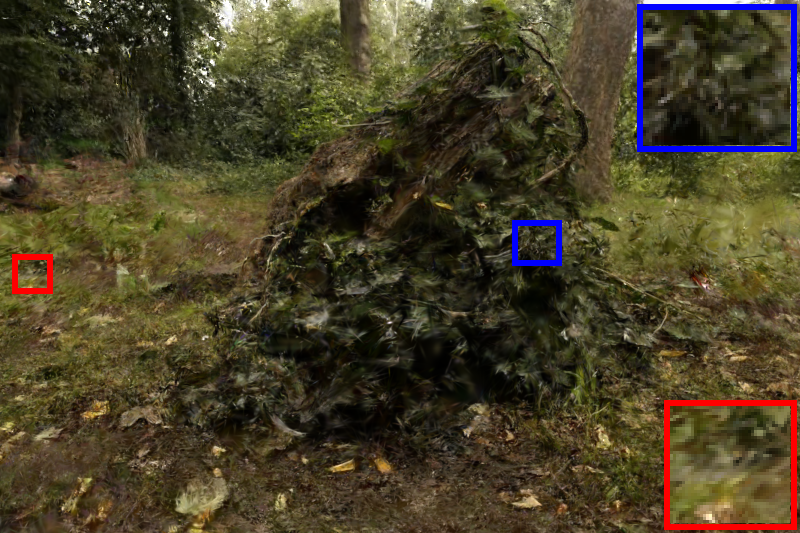} &
\includegraphics[width=\linewidth]{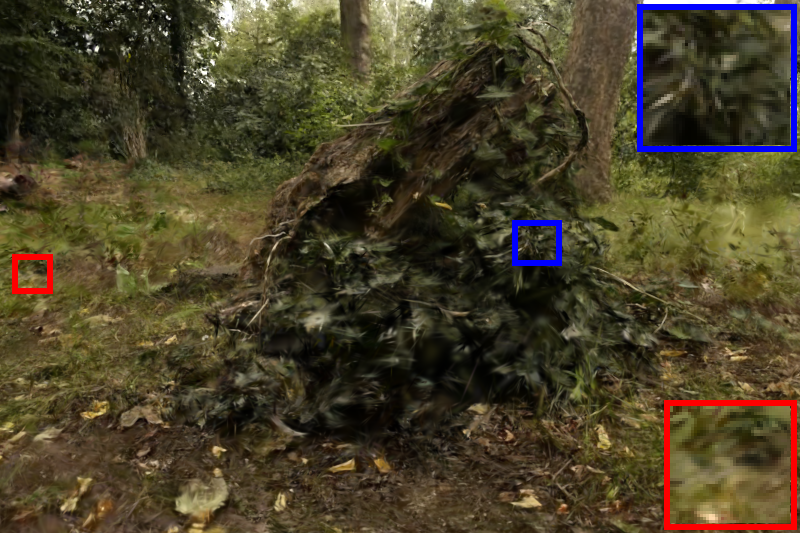} \\

\end{tabular}}%

\caption{\textbf{Qualitative result on Mip‑NeRF360.} \textit{From left to right are ground truth, Ours (\pname{}), FisheRF, and Random. Each row corresponds to a different scene. \textbf{20} views were selected to train a model and render the result on the test set. The blue box circles the object of our interest, while the red box circles some of the background.}}
  \label{fig:mip_transposed}
  \vspace{-3mm}
\end{figure*}

\noindent\textbf{Baselines.} We compare our approach against several state-of-the-art methods that explicitly model uncertainty for \textit{next-best-view} (NBV) selection. Our baseline pool includes: \emph{ActiveNeRF}~\cite{panActiveNeRFLearningWhere2022}, \emph{FisherRF}~\cite{jiangFisherRFActiveView2024}, and \emph{Bayes’ Rays}~\cite{goli2023}. Furthermore, to ensure a thorough comparison, we include \textbf{\emph{GauSS-MI}}~\cite{xieGauSSMIGaussianSplatting2025}, a recent information-theoretic method that represents the state-of-the-art in scene-level active reconstruction. We also note the concurrent work \textbf{\emph{POp-GS}}~\cite{wilsonPOpGSNextBest2025a}, which also focuses on NBV selection. However, as its implementation was not publicly available during our experimental phase, a direct quantitative comparison was not feasible. All baselines are trained using their official repositories with default hyper-parameters. To control for segmentation bias, we feed identical SAM-2 object masks to every method where applicable. NBV selection is executed with identical setting to ensure a fair comparison.
\begin{table*}[ht]
  \centering
\caption{\textbf{Validation of our parameter-centric FIM as a standalone uncertainty estimator.} \textit{To isolate the quality of our core uncertainty model, we evaluate it on the full scene without any object-aware masks. The table reports the Area Under the Sparsification Error (AUSE), a metric for uncertainty quality (lower is better).}}
  \scalebox{1}{
  \small
  \setlength{\tabcolsep}{4pt}
  \begin{tabular}{p{1.8cm}|cc|cc|cc|cc|cc|cc}
    \noalign{\vskip .1em}        
    \noalign{\hrule height .2em} 
    \noalign{\vskip .1em}     
    \multicolumn{1}{c|}{} &
    \multicolumn{2}{c|}{\textbf{africa}} &
    \multicolumn{2}{c|}{\textbf{basket}} &
    \multicolumn{2}{c|}{\textbf{statue}} &
    \multicolumn{2}{c|}{\textbf{torch}} &
    \multicolumn{2}{c|}{\textbf{TNT-Train}} &
    \multicolumn{2}{c}{\textbf{TNT-Truck}} \\
    
    \cmidrule[0.5pt](rl){2-3} \cmidrule[0.5pt](rl){4-5}
    \cmidrule[0.5pt](rl){6-7} \cmidrule[0.5pt](rl){8-9}
    \cmidrule[0.5pt](rl){10-11} \cmidrule[0.5pt](rl){12-13}
    
    & $\Delta$MSE$\downarrow$ & $\Delta$MAE$\downarrow$
    & $\Delta$MSE$\downarrow$ & $\Delta$MAE$\downarrow$
    & $\Delta$MSE$\downarrow$ & $\Delta$MAE$\downarrow$
    & $\Delta$MSE$\downarrow$ & $\Delta$MAE$\downarrow$
    & MSE$\downarrow$ & MAE$\downarrow$
    & MSE$\downarrow$ & MAE$\downarrow$ \\
    
    \toprule
    
    ActiveNeRF &
    1.123 & 0.958 & 0.642 & 0.546 & 0.818 & 0.732 & 1.513 & 1.246 &
    1.279 & 1.076 & 0.994 & 0.438 \\

    Bayes' Rays &
    \cellcolor{third}0.445 & \cellcolor{third}0.271 & \cellcolor{third}0.326 & \cellcolor{third}0.284 & \cellcolor{third}0.192 & \cellcolor{third}0.182 & \cellcolor{third}0.342 & \cellcolor{second}0.224 &
    \cellcolor{second}0.822 & \cellcolor{third}0.689 & \cellcolor{third}0.865 & \cellcolor{second}0.529 \\

    FisherRF &
    \cellcolor{best}0.181 & \cellcolor{best}0.186 & \cellcolor{second}0.212 & \cellcolor{second}0.225 &
    \cellcolor{second}0.191 & \cellcolor{best}0.178 & \cellcolor{best}0.247 & \cellcolor{third}0.254 &
    \cellcolor{third}0.892 & \cellcolor{second}0.632 & \cellcolor{second}0.843 & \cellcolor{third}0.589 \\
    
    \pname{} &
    \cellcolor{second}0.192 & \cellcolor{second}0.187 & \cellcolor{best}0.122 & \cellcolor{best}0.131 &
    \cellcolor{best}0.181 & \cellcolor{second}0.181 & \cellcolor{second}0.248 & \cellcolor{best}0.217 &
    \cellcolor{best}0.787 & \cellcolor{best}0.589 & \cellcolor{best}0.651 & \cellcolor{best}0.487 \\

    \noalign{\vskip .1em}        
    \noalign{\hrule height .2em} 
    \noalign{\vskip .1em}       
  \end{tabular}
  }
\vspace{-2mm}
  \label{tab:tnt_lf}
\end{table*}

\noindent\textbf{Metrics.} We evaluate reconstruction quality at two levels: \textbf{Scene-Level (Panoramic)} and \textbf{Object-Level}. Panoramic metrics, including Peak Signal-to-Noise Ratio (PSNR)~\cite{PSNR} for reconstruction accuracy, Structural Similarity Index (SSIM)~\cite{SSIM} for structural fidelity, and Learned Perceptual Image Patch Similarity (LPIPS)~\cite{LPIPS} for perceptual quality, are computed on the full, unmasked image to assess global consistency. Object-Level metrics are restricted strictly to the masked region of interest. 
Following the experimental protocol of~\cite{goli2023}, we calibrate our predictive uncertainty using the Area Under the Sparsification Error curve (AUSE). For every test image, we compute the per-pixel absolute error and the corresponding predicted uncertainty, then iteratively mask out the top $t\%$ of pixels ($t = 1,\dots,100$) in descending order of predicted uncertainty. Integrating the resulting error curve yields $\text{AUSE}_{\text{MAE}}$ and $\text{AUSE}_{\text{MSE}}$, where lower values indicate better alignment between predicted uncertainty and actual reconstruction error.

\noindent\textbf{Implementation Details.}
Following SoTA NBV protocols~\cite{wilsonPOpGSNextBest2025a}, we evaluate on the benchmark datasets described in Sec.~\ref{sec:Dataset}. The Gaussians are initialised with COLMAP~\cite{Schonberger_2016_CVPR}, and object masks are obtained from SAM-2~\cite{ravi2024sam2segmentimages}. NBV planning follows FisherRF~\cite{jiangFisherRFActiveView2024}: four initial views are selected using the farthest-point strategy, followed by 100 epochs of training and the
addition of a new view chosen by the highest predicted uncertainty
within the object mask. The selection process repeats until 20 views are reached, after which the model is optimised for 30k iterations with the default 3DGS schedule.
To ensure practical scalability, \pname{} incorporates strategic optimizations to minimize the computational overhead of view selection. 
The main overhead arises during NBV planning, where we evaluate the explicit Jacobian projection $J_u=\partial C(u)/\partial\theta$ for candidate views (Eq.~\ref{eq:final_uncertainty}); computing this densely for every pixel at full resolution would be prohibitive. We therefore combine a diagonal FIM approximation, which reduces covariance storage and propagation from $O(d^2)$ to $O(d)$, with a strided patch-based sampling scheme during planning: we score candidate views strictly within the masked region of interest using uniformly sampled $16{\times}16$ patches at a stride of 16 pixels. This strategy ignores vast background areas and reduces the evaluated pixel count by $\sim256\times$ (relative to the mask) while retaining sufficient spatial coverage for reliable ranking. These choices keep uncertainty tracking lightweight during training and make Jacobian-based projection practical at selection time, enabling scalability to large scenes (validated on TNT sequences with $>1.2$M Gaussians).
\vspace{-3mm}
\subsection{Quantitative Results}
\label{OverallRenderingQuality}
Our primary quantitative evaluation, presented in table~\ref{tab:full}, analyzes reconstruction quality at two levels of granularity: the full scene and the object of interest.

\textbf{Scene-Level Panoramic Performance.}
In the panoramic evaluation, which assesses the entire rendered view, GauSS-MI demonstrates strong performance due to its entropy-based global objective. OUGS remains highly competitive, consistently outperforming FisherRF and ranking second only to GauSS-MI in our benchmarks. This reflects an intentional design choice: we prioritize robust global consistency to prevent catastrophic background degradation, while strategically reserving the active selection budget for the target object. Notably, panoramic metrics are computed on the full image without masking; moreover, all methods receive identical masks during view selection, so improvements over FisherRF cannot be attributed to privileged semantic access.

\textbf{Object-Aware Performance.}
The decisive advantage of our framework is revealed in the object-centric evaluation. In this critical setting, \textbf{\pname{} consistently and substantially outperforms all baselines, including GauSS-MI, across every dataset, table~\ref{tab:tnt_lf}}.
Crucially, employing \textbf{identical SAM-2 masks for all baselines} eliminates segmentation quality as a confounding variable.
The results validate our core hypothesis: while scene-level methods dilute their view budget on high-gradient background textures, our method successfully \textbf{reallocates the finite view budget} to the target, achieving significantly higher fidelity on the regions of interest.

\vspace{-2mm}
\subsection{Qualitative Results}
\label{QualitativeResults}
\begin{figure*}[t]
  \centering
  \setlength{\tabcolsep}{0.3pt}
  \renewcommand{\arraystretch}{0.3}
  \resizebox{\textwidth}{!}{%
    \begin{tabular}{@{}lccc@{}}
      & GT & Render & Uncertainty \\
      & 
        \includegraphics[width=0.32\linewidth]{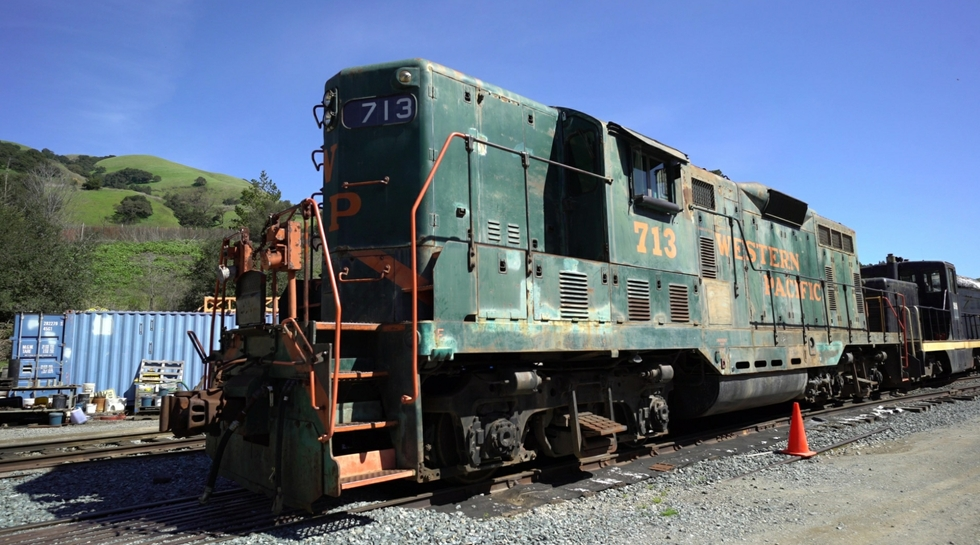} &
        \includegraphics[width=0.32\linewidth]{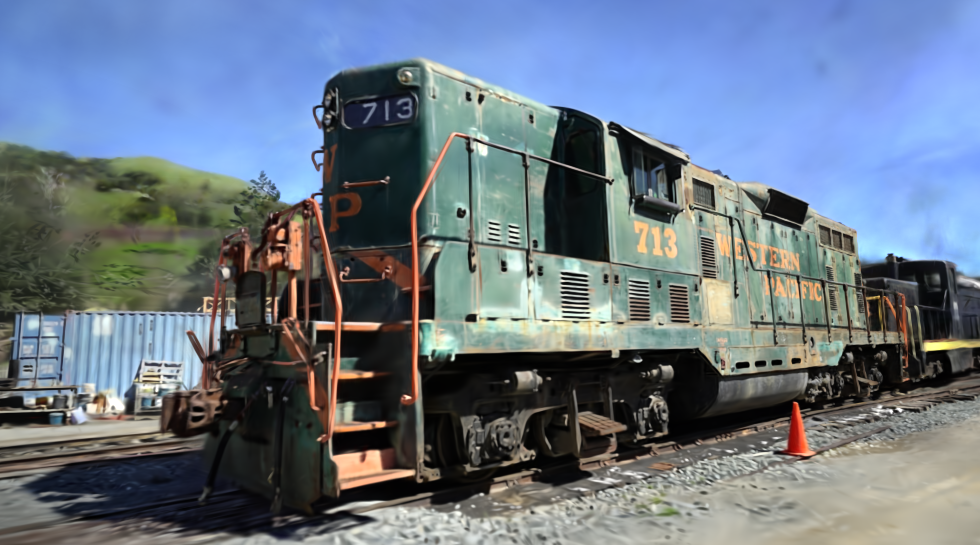} &
        \includegraphics[width=0.32\linewidth]{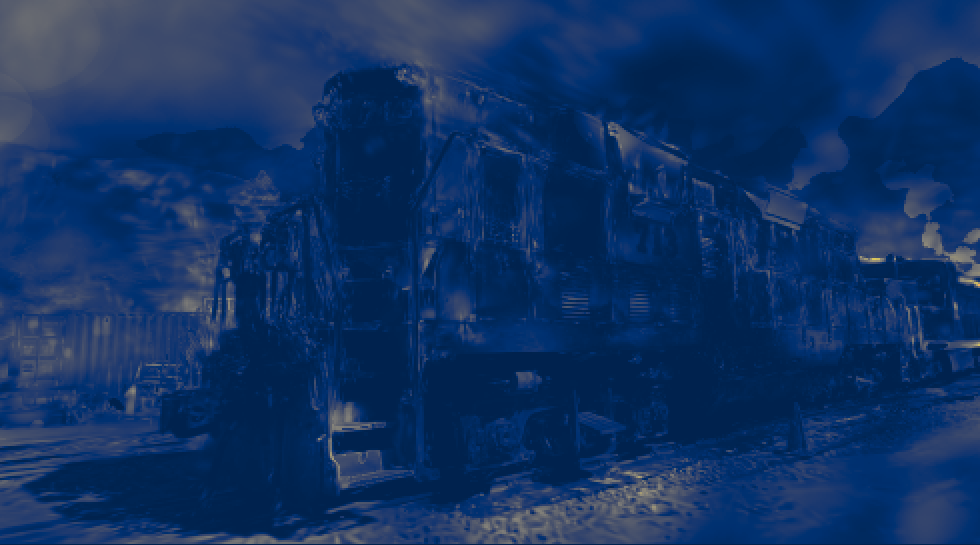} \\
      &
        \includegraphics[width=0.32\linewidth]{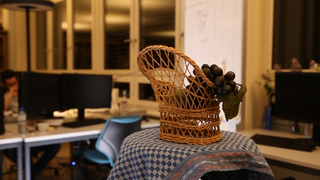} &
        \includegraphics[width=0.32\linewidth]{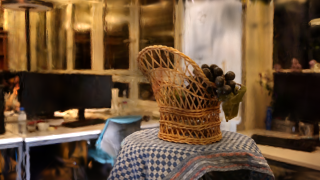} &
        \includegraphics[width=0.32\linewidth]{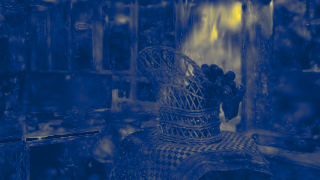}
    \end{tabular}%
  }

\caption{\textbf{Qualitative validation of our parameter-centric uncertainty.}\textit{ These results on the TNT-Train and LF-Basket scenes showcase the remarkable accuracy of our uncertainty estimation. The uncertainty heatmap (right, yellow indicates high uncertainty) precisely localizes the regions where the final rendering (middle) deviates from the ground truth (left), such as blurry structures and ghosting artifacts. This strong visual correlation demonstrates the effectiveness of our physically-grounded model in predicting and explaining rendering errors.}}
\vspace{-4mm}
  \label{fig:TNT-uc}
\end{figure*}

Figure~\ref{fig:obj_active_view} illustrates the convergence behavior, comparing our approach against the FisherRF. While both methods track closely in the panoramic evaluation, OUGS achieves sharp gains in PSNR and SSIM early in the process by prioritizing object centric views, whereas FisherRF’s progress on the object is slower due to distractions from background gradients. In addition, Figure~\ref{fig:mip_transposed} provides a qualitative comparison on the Mip-NeRF 360 dataset. To specifically analyze the behavior of uncertainty estimation based on the Fisher Information Matrix, we focus the visual comparison on our method, FisherRF, and a random baseline.

The results clearly illustrate the impact of our object-aware strategy. Across all scenes, our reconstructions remain visually closest to the ground truth in the regions of interest (blue boxes). On the \textit{stump} scene, for example, the slender bark fibres are sharply delineated in our result, whereas they are rendered as blurred or are entirely missing by FisherRF. This reveals the fundamental difference in FIM-based approaches: FisherRF, however, often produces a cleaner background (red boxes). This stems from its scene-level Fisher information score; high-gradient background textures can dominate the score and steer the next-best-view search away from the object. Our method, by design, resists this distraction. Random sampling, while uninformed, spreads views uniformly and therefore sometimes captures object-centric angles, leading to occasional details that surpass FisherRF, but this comes at the cost of high variance and no guarantees. The visual evidence strongly indicates that our method preserves object detail most faithfully. The modest artifacts that may appear at the image periphery do not outweigh the substantial and consistent gains in the target region, confirming the effectiveness of our object-aware formulation.

\subsection{Ablation Study}
\label{sec:ablation}

In this section, we conduct a rigorous analysis to isolate the sources of our performance gains, verify robustness against segmentation failures, and quantify EMA update strategy.

\textbf{Quantitative Validation. 
As reported in table~\ref{tab:tnt_lf}, OUGS demonstrates robust uncertainty calibration. While performing \textbf{on par} with FisherRF~\cite{jiangFisherRFActiveView2024} on simpler, foreground-centric scenes (e.g., \textit{Africa}, \textit{Torch}), OUGS achieves \textbf{superior} calibration on geometrically complex environments. 
Notably, on the challenging \textit{TNT-Train} and \textit{TNT-Truck} scenes, our method reduces AUSE significantly compared to the baselines.
This confirms that our parameter-to-pixel Jacobian formulation provides a more accurate proxy for reconstruction error, particularly in complex scenarios. The semantic mask therefore acts as a spatial filter, but the high-quality uncertainty signal is intrinsic to our FIM formulation. }

\textbf{Qualitative Visualization.} Figure~\ref{fig:TNT-uc} complements these findings. The uncertainty highlighted by our model precisely concentrates on background regions that later exhibit blur or ghosting artifacts, while high-confidence areas remain artifact-free. This visual alignment confirms that our parameter-level uncertainty estimation successfully localizes residual errors in image space.

\begin{figure*}[t]
  \centering
  \includegraphics[width=0.95\linewidth]{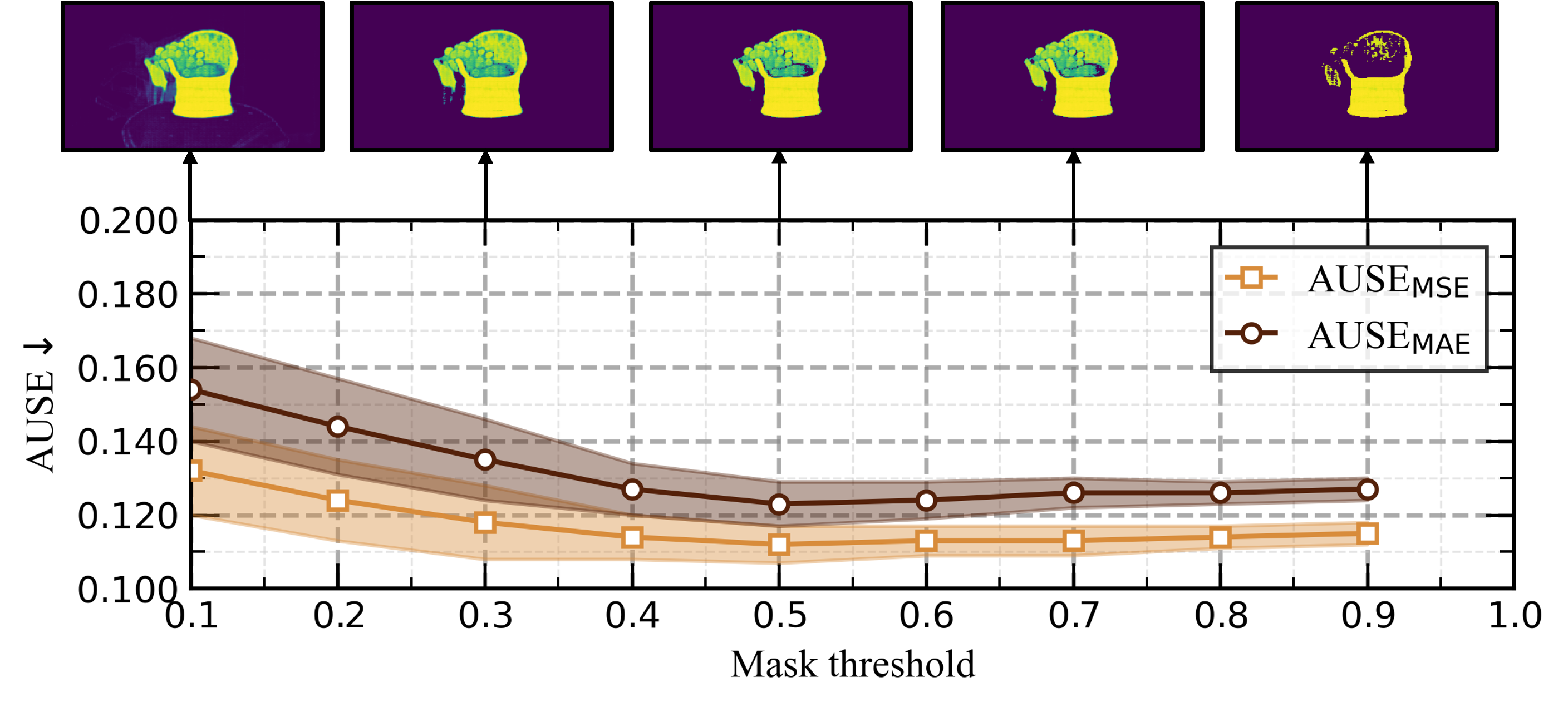}
  \vspace{-2mm}
  \caption{\textbf{Sensitivity to mask quality.} \textit{We analyze object-level AUSE (lower is better) by varying the mask binarization threshold. The clear basin of stability indicates that OUGS is robust to a wide range of segmentation hyperparameters and effectively utilizes the mask to filter background clutter without requiring pixel-perfect thresholds.}}
  \label{fig:ablation}
  \vspace{-4mm}
\end{figure*}
\vspace{-2mm}

\textbf{Source of Gains: Intrinsic Uncertainty Quality}
A core question raised in our analysis is whether our gains stem solely from object masking or from a fundamental improvement in uncertainty estimation. To isolate the quality of our core uncertainty model from semantic guidance, we refer to the \textbf{Area Under the Sparsification Error (AUSE)} evaluated on the \textbf{full scene} without applying any object masks in table~\ref{tab:tnt_lf}.

\textbf{Robustness to Imperfect Segmentation}
\label{sec:ablation-mask}
Since our method leverages semantic masks for object targeting, a critical question is how robust \pname{} is to segmentation errors. We analyze this in two dimensions: sensitivity to thresholding parameters and resilience to severe segmentation noise.

\textbf{Sensitivity to Probability Thresholds.}
To investigate how the quality of the soft mask influences our \mbox{object‑aware} uncertainty estimation, we simulate mask degradation by varying the threshold value from 0.1 to 0.9 and plot the corresponding object level AUSE scores. Pixels with probabilities above the threshold are considered part of the object. At low threshold values (left side of the plot), the mask includes more unwanted background regions, leading to suboptimal view selection and higher (worse) AUSE scores. As the threshold increases, the mask becomes more focused on the object, improving performance and reaching an optimal AUSE around 0.5—where the mask best isolates the object while excluding background clutter. Beyond this point, further increases in the threshold aggressively remove less salient object regions, slightly degrading performance as informative areas are excluded.

\textbf{Robustness to Segmentation Noise.}
To evaluate robustness under realistic segmentation failures, we stress-test both the \emph{view-selection policy} and the \emph{final object reconstruction} on \textit{TNT--Truck} by injecting three controlled perturbations into the planning masks: \textbf{(i) Under-segmentation}, which removes portions of the true foreground; \textbf{(ii) Over-segmentation}, which introduces spurious foreground blobs in the background; and \textbf{(iii) Boundary Blur}, which dilates the mask to include surrounding clutter.

\begin{figure}[h]
    \centering
    \includegraphics[width=1.0\linewidth]{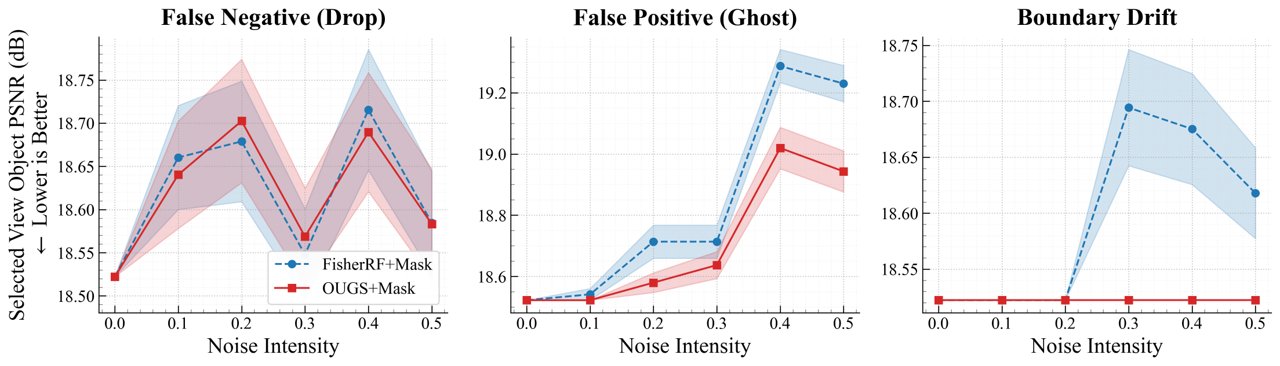}
    \vspace{-2mm}
    \caption{\textbf{NBV robustness under segmentation noise (TNT--Truck).} \textit{Selected-view object PSNR (lower is better) versus noise intensity for Drop (left), Ghost (center), and Boundary Drift (right).}}
    \label{fig:mask_robustness}
    \vspace{-2mm}
\end{figure}

Figure~\ref{fig:mask_robustness} analyzes one-step selection robustness via \emph{Selected View Object PSNR} (\textit{lower is better}), as an effective NBV policy should select views where the object is currently worst reconstructed. Under \textbf{Under-segmentation} (left), both methods degrade similarly—missing foreground pixels reduce evidence for both scoring rules. Under \textbf{Over-segmentation} (center), the baseline increasingly selects already well-reconstructed views (higher PSNR), wasting view budget on background artifacts. Under \textbf{Boundary Blur} (right), this gap widens: as masks expand into clutter, the baseline shifts toward ``easy'' views while \pname{} remains stable, demonstrating robustness to coarse boundary errors.

\begin{table}[t]
\centering
\renewcommand{\arraystretch}{1.08}
\setlength{\tabcolsep}{6pt}
\vspace{-1mm}
\small
\caption{\textbf{Robustness of object-aware reconstruction on TNT under degraded object masks during NBV planning.} }
\label{tab:mask_robustness}

\resizebox{0.95\linewidth}{!}{
\begin{tabular}{l|cccc}
\toprule
Method & PSNR$\uparrow$ & PSNR Drop$\downarrow$ & SSIM$\uparrow$ & LPIPS$\downarrow$ \\
\midrule
\multicolumn{5}{l}{\textit{I. Clean masks (ideal)}} \\
\hline
FisherRF & 23.48 & \textemdash & 0.921 & 0.140 \\
\textbf{OUGS (ours)} & \textbf{26.25} & \textemdash & \textbf{0.933} & \textbf{0.117} \\
\midrule
\multicolumn{5}{l}{\textit{II. ``Ghost'' noise (false positives)}} \\
\hline
FisherRF & 20.15 & \textcolor{red}{3.33} & 0.845 & 0.210 \\
\textbf{OUGS (ours)} & \textbf{24.05} & \textcolor{teal}{\textbf{2.20}} & \textbf{0.895} & \textbf{0.152} \\
\midrule
\multicolumn{5}{l}{\textit{III. ``Boundary drift'' (coarse edges)}} \\
\hline
FisherRF & 21.60 & \textcolor{red}{1.88} & 0.872 & 0.185 \\
\textbf{OUGS (ours)} & \textbf{24.45} & \textcolor{teal}{\textbf{1.80}} & \textbf{0.902} & \textbf{0.141} \\
\bottomrule
\end{tabular}
}
\vspace{-5mm}
\end{table}
\definecolor{best}{rgb}{1, 0.7, 0.7}
\definecolor{second}{rgb}{1, 0.85, 0.7}
\definecolor{third}{rgb}{1, 1, 0.7}

\begin{table*}[t]
  \centering

\rev{\caption{\textbf{Quantitative comparison across four scenes of the LF dataset.} \textit{We compare our Warp-based mask generation against ground-truth SAM2 masks (GT). Metrics are evaluated on object regions after training with 20 views for 30k iterations.}} \label{tab:mask_comparison}}
  \scalebox{1}{
  \small
  \setlength{\tabcolsep}{3.5pt}
  \begin{tabular}{l|ccc|ccc|ccc|ccc}
    \toprule
    \multicolumn{1}{c|}{} &
    \multicolumn{3}{c|}{\textbf{africa}} &
    \multicolumn{3}{c|}{\textbf{basket}} &
    \multicolumn{3}{c|}{\textbf{statue}} &
    \multicolumn{3}{c}{\textbf{torch}} \\
    \cmidrule[0.5pt](rl){2-4} \cmidrule[0.5pt](rl){5-7}
    \cmidrule[0.5pt](rl){8-10} \cmidrule[0.5pt](rl){11-13}
    & PSNR$\uparrow$ & SSIM$\uparrow$ & LPIPS$\downarrow$
    & PSNR$\uparrow$ & SSIM$\uparrow$ & LPIPS$\downarrow$
    & PSNR$\uparrow$ & SSIM$\uparrow$ & LPIPS$\downarrow$
    & PSNR$\uparrow$ & SSIM$\uparrow$ & LPIPS$\downarrow$ \\
    \midrule

    \textbf{Online FisherRF} &
    24.80 & 0.975 & 0.018 &
    24.95 & 0.958 & 0.048 &
    27.60 & 0.972 & 0.032 &
    22.60 & 0.958 & 0.029 \\

    \textbf{Offline FisherRF} &
    \cellcolor{second}32.45 & \cellcolor{second}0.991 & \cellcolor{second}0.011 &
    \cellcolor{second}29.35 & \cellcolor{second}0.973 & \cellcolor{second}0.038 &
    \cellcolor{second}31.85 & \cellcolor{second}0.983 & \cellcolor{third}0.024 &
    \cellcolor{second}29.66 & \cellcolor{second}0.978 & \cellcolor{second}0.020 \\

    \textbf{Online OUGS} &
    \cellcolor{third}27.94 & \cellcolor{third}0.988 & \cellcolor{third}0.014 &
    \cellcolor{third}27.35 & \cellcolor{third}0.968 & \cellcolor{third}0.041 &
    \cellcolor{third}30.02 & \cellcolor{third}0.981 & \cellcolor{second}0.023 &
    \cellcolor{third}25.63 & \cellcolor{third}0.969 & \cellcolor{third}0.025 \\
    
    \textbf{Offline OUGS} &
    \cellcolor{best}33.82 & \cellcolor{best}0.996 & \cellcolor{best}0.010 &
    \cellcolor{best}30.65 & \cellcolor{best}0.982 & \cellcolor{best}0.036 &
    \cellcolor{best}33.25 & \cellcolor{best}0.989 & \cellcolor{best}0.022 &
    \cellcolor{best}31.02 & \cellcolor{best}0.988 & \cellcolor{best}0.019 \\
    
    \bottomrule
    
  \end{tabular}
  }
  \vspace{-4mm}
\end{table*}

Table~\ref{tab:mask_robustness} confirms this selection stability translates to end-to-end robustness. Under \textbf{Over-segmentation}, FisherRF drops 3.33,dB by overfitting to background artifacts, while \pname{} degrades only 2.20,dB. This resilience stems from Eq.~\ref{eq:final_uncertainty}: our uncertainty score combines semantic mask weights with physical uncertainty from the Jacobian. False positive regions typically have low geometric uncertainty, suppressing their contribution and acting as a physics-based safeguard against mask noise.

\textbf{Reliability in Fully Autonomous Settings.}
A key challenge in object-centric reconstruction is obtaining object masks for unseen candidate views. Beyond offline SAM2-based propagation, we evaluate an online depth-guided warping strategy that uses the current 3DGS geometry to transfer masks to unvisited viewpoints, but its quality is limited by view coverage and depth fidelity. Table~\ref{tab:mask_comparison} shows that this online variant underperforms the SAM2 baseline; the gap is most pronounced at the beginning of scanning, where limited view coverage yields noisier depth-warped masks and consequently less effective NBV decisions. Although mask quality improves over NBV iterations as coverage increases, the view budget consumed early cannot be recovered, leading to lower final reconstruction quality. This trend aligns with Table~\ref{tab:mask_robustness}, where degraded masks (e.g., projection-induced boundary drift due to geometric errors) measurably harm planning; we therefore recommend high-quality mask acquisition (e.g., SAM2) whenever available.

\subsubsection*{Analysis of the EMA Update Schedule} 
Our FIM approximation relies on an online update of the diagonal entries using an EMA of squared gradients. A key design choice is our decaying momentum schedule for the EMA parameter, $\alpha_t = 0.95^{\,1.5t/T}$, which prioritizes stability early in training and faster adaptation later on. To validate this choice, we conduct an ablation study comparing our proposed schedule against simpler alternatives with a constant momentum. We evaluate the final object-aware PSNR on the LF scene after 20 selected views.

\begin{table}[h]
\centering
\caption{\textbf{Ablation on the EMA update schedule, evaluated on the LF-Statue scene.} \textit{Our decaying momentum strategy outperforms all constant momentum alternatives.}}
\small
\vspace{-2mm}

\begin{tabular}{lc}
\toprule
\textbf{EMA Schedule Strategy} & \textbf{Object-Aware PSNR (dB)} $\uparrow$ \\
\midrule
No EMA (Instantaneous) & 29.52 \\
Low Momentum ($\alpha_t=0.90$) & 31.63 \\
Medium Momentum ($\alpha_t=0.95$) & 31.91 \\
High Momentum ($\alpha_t=0.99$) & 32.04 \\
\midrule
\rowcolor{gray!20}
\textbf{Ours (Decaying Momentum)} & \textbf{32.18} \\
\bottomrule
\end{tabular}
\vspace{-1mm}

\label{tab:ema_ablation}
\end{table}

The results, presented in table~\ref{tab:ema_ablation}, confirm the effectiveness of our proposed strategy. A constant high momentum ($\alpha_t=0.99$) is overly cautious, smoothing too much and preventing the model from adapting quickly enough, resulting in lower PSNR. Conversely, a constant low momentum ($\alpha_t=0.9$) is too reactive to noisy gradients, leading to an unstable FIM estimate and the worst performance. Our decaying schedule strikes the optimal balance, achieving the highest PSNR. This validates that our carefully designed "slow-start, fast-finish" update strategy is crucial for robustly estimating the FIM online and contributes significantly to the final reconstruction quality.
\vspace{-2mm}

\vspace{-2mm}
\section{Limitations and Future Work}
\label{sec:limitations}

While our framework demonstrates a significant advancement in object-centric active reconstruction, we acknowledge several limitations that open up exciting avenues for future research. 
We employ SAM-2 as an offline oracle to simulate reliable masks, decoupling planning efficacy from segmentation failures.
Consequently, our method's effectiveness is contingent upon the availability and quality of a semantic mask for the object of interest. While our ablation study (Sec.~\ref{sec:ablation-mask}) demonstrates robustness to moderate mask degradation, significant segmentation errors inevitably degrade reconstruction quality; we observe that severe noise such as spurious Ghost regions can cause performance drops comparable to baselines. Additionally, while we extend this method to handle online mask estimation (e.g., via depth image warping), achieving high-fidelity reconstruction without oracle masks still remains a quality challenge. While we present a theoretical extension to multiple objects in appendix, which is further detailed in the Supplementary Material, the experimental evaluation in this work focuses on single-object scenarios; empirical validation of multi-object active reconstruction and optimal weighting strategies remain to be explored. Furthermore, our method approximates the full Fisher Information Matrix with its diagonal entries, assuming independence between different parameters of a Gaussian—a strong simplification that could be relaxed with more structured FIM approximations.

\vspace{-3mm}
\section{Conclusion}
We introduced \pname{}, an object-aware uncertainty estimation framework for active view selection in 3DGS. Our work presents a fundamental shift in how uncertainty is modeled for explicit representations. By deriving uncertainty directly from the physical parameters of the 3D Gaussian primitives, we establish a more principled and interpretable link between the 3D scene representation and the rendered 2D image.
Our method projects this parameter-level covariance into pixel space and, by coupling it with semantic masks, produces an uncertainty score that effectively disentangles the object of interest from its environment. This is enabled by an efficient diagonal FIM update scheme that makes the approach computationally tractable. When integrated into a next-best-view loop, our method consistently and substantially outperforms state-of-the-art baselines in the critical task of object-centric reconstruction, achieving sharper results under the same view budget. Notably, our underlying uncertainty model also proves to be highly competitive for global scene reconstruction. Ultimately, these results underscore the importance of disentangling object-level uncertainty from background clutter for efficient, high-fidelity active reconstruction.
\appendix
\clearpage
\appendix

\section{Jacobian--Covariance Propagation}
\label{sec:jacob_propagation}

Let $C(u;\theta)\in\mathbb R^{3}$ denote the rendered RGB colour at pixel $u$
under parameters $\theta\in\mathbb R^{d}$, and let $\theta^\star$ be the MAP
estimate after optimisation. We model local parameter uncertainty via
$\delta\theta:=\theta-\theta^\star$ with mean $\mathbb E[\delta\theta]=0$
and covariance $\Sigma=\mathbb E[\delta\theta\,\delta\theta^{\top}]$.

\paragraph{Second-order expansion.}
A Taylor expansion of $C(u;\theta)$ around $\theta^\star$ yields
\begin{equation}
\label{eq:taylor_app}
C(u;\theta)
=
C(u;\theta^\star)
+
J_u\,\delta\theta
+
\frac12\,\delta\theta^{\top}H_u\,\delta\theta
+\mathcal O(\|\delta\theta\|^3),
\end{equation}
where
\[
J_u := \left.\frac{\partial C(u;\theta)}{\partial\theta}\right|_{\theta^\star}
\in\mathbb R^{3\times d} \quad\text{and}\quad H_u := \left.\frac{\partial^2 C(u;\theta)}{\partial\theta^2}\right|_{\theta^\star}
\in\mathbb R^{d\times d}.
\]

\paragraph{First-order (Jacobian) covariance.}
For uncertainty propagation we use the standard first-order approximation,
which is accurate when $\delta\theta$ is small (equivalently, when $\Sigma$ is
small in operator/Frobenius norm). Retaining only the linear term in
\eqref{eq:taylor_app} gives
\begin{equation}
\label{eq:lin_app}
C(u;\theta)\;\approx\;C(u;\theta^\star)+J_u\,\delta\theta.
\end{equation}
Taking the second central moment yields the classic Jacobian--covariance law:
\begin{equation}
\label{eq:jacprop}
\operatorname{Var}\!\bigl[C(u;\theta)\bigr]
\;\approx\;
J_u\,\Sigma\,J_u^{\!\top}.
\end{equation}
Intuitively, $J_u$ acts as a sensitivity map: its element
\(
[J_u]_{kl}=\partial C_k(u;\theta^\star)/\partial\theta_l
\)
quantifies how the $l$-th parameter perturbs the $k$-th colour channel at pixel
$u$, and \eqref{eq:jacprop} weights these sensitivities by parameter variance.

\paragraph{Remainder control.}
Including the quadratic term in \eqref{eq:taylor_app} contributes corrections of
higher order in $\Sigma$. Under mild smoothness, one obtains a bound of the form
\begin{equation}
\label{eq:remainder_app}
\bigl\|\operatorname{Var}[C(u;\theta)]-J_u\Sigma J_u^{\top}\bigr\|
\;\le\;
c_u\,\|\Sigma\|^2
\;+\;\mathcal O(\|\Sigma\|^3),
\end{equation}
 Hence the
approximation error vanishes as the Gaussian parameters become well-constrained.
\vspace{-4mm}
\section{FIM Approximation and Online Update}
\label{sec:fisher_appendix}

\textbf{Fisher Matrix under Gaussian Image Noise.} Assume independent Gaussian image noise
$\varepsilon(u)\sim\mathcal N(0,\sigma^{2}\mathbf I_{3})$ so that
$C_{\mathrm{gt}}(u)=C(u;\theta)+\varepsilon(u)$.
The (per-iteration) negative log-likelihood is the photometric error
\begin{equation}
\label{eq:ell_app}
\ell_t(\theta)
=\frac1{2\sigma^{2}}
\sum_{u\in\Omega_t}
\bigl\|C_{\mathrm{pred},t}(u;\theta)-C_{\mathrm{gt},t}(u)\bigr\|_{2}^{2},
\end{equation}
where $\Omega_t$ denotes the set of sampled pixels (or patches) at step $t$.
Let $r_t(u):=C_{\mathrm{pred},t}(u;\theta)-C_{\mathrm{gt},t}(u)$ be the residual.
Then
\begin{equation}
\label{eq:grad_app}
\nabla_{\theta}\ell_t
=
\frac1{\sigma^{2}}
\sum_{u\in\Omega_t}
J_u^{\top}r_t(u).
\end{equation}

The Fisher information matrix (FIM) is defined as the noise expectation of the
outer product of the log-likelihood gradients:
\begin{equation}
\label{eq:fim_def_app}
\mathcal I
=
\mathbb E_{\varepsilon}\!\bigl[\nabla_{\theta}\ell_t\;\nabla_{\theta}\ell_t^{\top}\bigr].
\end{equation}
Under the i.i.d.\ Gaussian noise assumption, cross-terms vanish and
$\mathbb E[r_t(u)r_t(u)^{\top}]=\sigma^2\mathbf I_3$, yielding
\begin{equation}
\label{eq:fim_sum_app}
\mathcal I
=
\frac1{\sigma^{2}}
\sum_{u\in\Omega_t}
J_u^{\top}J_u.
\end{equation}
Equivalently, stacking per-pixel Jacobians row-wise into
$J\in\mathbb R^{(3|\Omega_t|)\times d}$ gives $\mathcal I=\sigma^{-2}J^{\top}J$.
This dense $d{\times}d$ matrix captures parameter correlations but is
prohibitively expensive to store or invert at scale.

\textbf{Diagonal Approximation and Damping.} A local Laplace / Cram\'er--Rao approximation gives the posterior covariance
\begin{equation}
\label{eq:laplace_app}
\Sigma
\;\approx\;
\sigma^{2}\bigl(\mathcal I+\lambda\mathbf I_d\bigr)^{-1},
\end{equation}
where $\lambda>0$ is a small Tikhonov damping term for numerical stability.
To obtain a tractable estimate, we retain only the diagonal Fisher entries:
\begin{equation}
\label{eq:fim_diag_app}
\mathcal I^{\mathrm{diag}}
=
\operatorname{diag}(\mathcal I)
=
\frac1{\sigma^{2}}\,
\operatorname{diag}\!\Bigl(\sum_{u\in\Omega_t}J_u^{\top}J_u\Bigr),
\qquad
\Sigma\approx\sigma^{2}\bigl(\mathcal I^{\mathrm{diag}}+\lambda\mathbf I_d\bigr)^{-1}.
\end{equation}
This approximation discards off-diagonal correlations while preserving
parameter-wise uncertainty magnitudes, which is sufficient for our
pixel-space propagation through $J_u$.

\textbf{EMA Update and Object-Aware Pixel Covariance.} During optimisation we maintain an online estimate of the diagonal Fisher via an
exponential moving average (EMA) of squared gradients:
\begin{equation}
\label{eq:fisher_ema_app}
\mathcal I_{t,i}^{(j)}
=
\alpha_t\,\mathcal I_{t-1,i}^{(j)}
+
(1-\alpha_t)\bigl[\nabla_{\theta_i^{(j)}}\ell_t\bigr]^2,
\qquad
\alpha_t = 0.95^{\,1.5\,t/T}.
\end{equation}
Here $\mathcal I_{t,i}^{(j)}$ denotes the $(j,j)$ entry (diagonal element) for
the $j$-th parameter of Gaussian $i$, and $T$ is the total number of optimisation
steps. The schedule starts at $\alpha_0=1$ and decays smoothly, gradually
increasing the contribution of new gradient information as optimisation
stabilises. With the diagonal Fisher approximation, the pixel-space covariance follows from
\eqref{eq:jacprop}. Introducing a soft object mask $M_k(u)\in[0,1]$ yields the
object-aware covariance used in the main paper:
\begin{equation}
\label{eq:object_cov_app}
\Sigma_{C,k}(u)
=
(M_k(u))^{2}\,
J_u\Bigl(\sigma^{2}\bigl(\mathcal I^{\mathrm{diag}}_{t}+\lambda\mathbf I_d\bigr)^{-1}\Bigr)
J_u^{\top}.
\end{equation}
Although $\mathcal I^{\mathrm{diag}}_{t}$ is diagonal, $J_u$ generally has
nonzero entries across many parameters, so the projection couples parameter
effects when forming pixel-space uncertainty. Summing
$\operatorname{tr}(\Sigma_{C,k}(u))$ over pixels with $M_k(u)>0$ produces an
object-centric uncertainty score that drives next-best-view selection.
\vspace{-4mm}
\section{Multi-Object Extension}
\label{sec:multi_object_extension}
While this paper focuses on single-object active reconstruction, the framework naturally extends to multiple targets. For $K$ objects with masks $M_k(u)$ and covariances $\Sigma_{C,k}(u)$, the aggregated score is:
\begin{equation}
\text{Score}(v) = \sum_{k=1}^{K} w_k \cdot \text{Trace}\left(\int_{\Omega} \Sigma_{C,k}(u) \, du\right),
\label{eq:multi_object_suppl}
\end{equation}
where weights $w_k$ satisfy $\sum_k w_k = 1$. The computational complexity remains $\mathcal O(K \cdot d)$ under diagonal FIM approximation. Determining optimal weighting strategies for specific applications (e.g., task-specific priorities, adaptive updates) represents an interesting direction for future work.
\clearpage
\bibliographystyle{eg-alpha-doi} 
\bibliography{main}      

\end{document}